\documentclass{article}

\usepackage{iclr2027_conference,times}
\usepackage{amsmath}
\usepackage{amssymb}
\usepackage{xcolor}
\usepackage{colortbl}
\usepackage{multirow}
\usepackage{hyperref}
\usepackage{url}
\usepackage{graphicx}
\usepackage{booktabs}
\usepackage{placeins}
\usepackage{float}
\usepackage[ruled,vlined]{algorithm2e}
\usepackage{wrapfig}

\title{MomWorld: Momentum-Aware Latent World Model for Long-Horizon Autonomous Driving}
\author{
\normalfont Ziying Song$^{1}$ \quad Shengkai Zhang$^{2}$ \quad Lei Yang$^{1}$ \quad Haozhuang Chi$^{1}$\\
Yuchen Liu$^{3}$ \quad Jiangtao Su$^{1}$ \quad Lin Liu$^{4}$ \quad Ziyang Liu$^{5}$ \quad Chen Lv$^{1}$\thanks{Corresponding author.}\\
\textsuperscript{1}Nanyang Technological University \quad \textsuperscript{2}Beijing Jiaotong University\\
\textsuperscript{3}North University of China \quad \textsuperscript{4}Dalian University of Technology \quad \textsuperscript{5}Tsinghua University\\
\url{https://github.com/modaxiansheng/MomWorld}
}

\iclrfinalcopy

\begin{document}
\maketitle

\begin{abstract}
Long-horizon planning enables autonomous vehicles to anticipate scene evolution and potential risks, supporting safe and stable decisions in complex interactions. However, existing methods struggle to propagate motion trends from observed history into the future. Long rollouts based on a single latent state may further attenuate useful dynamics, retain stale motion patterns, and disrupt reliable near-term plans. We introduce \textbf{MomWorld}, a momentum-aware latent world model for long-horizon planning. MomWorld extracts scene motion trends from historical-to-current observations and propagates latent momentum into future horizons, jointly predicting future configuration and momentum states. A learnable momentum persistence mechanism preserves stable trends, scene-conditioned momentum updates adapt future dynamics, and a scene-adaptive reset gate suppresses stale momentum under abrupt changes. We further propose MoFlow, a momentum-conditioned flow-matching module that refines a base trajectory to align with the predicted future scene evolution in only a few integration steps, with a horizon-aware residual fusion that preserves near-term planning stability while permitting stronger long-range corrections. Extensive experiments on NAVSIM, nuScenes and Bench2Drive demonstrate that MomWorld improves long-horizon planning consistency and reduces the average collision rate by 12.2\% relative to MomAD over a 6-second planning horizon.

\end{abstract}

\section{Introduction}

\begin{figure}[t]
    \centering
    \includegraphics[width=0.98\linewidth]{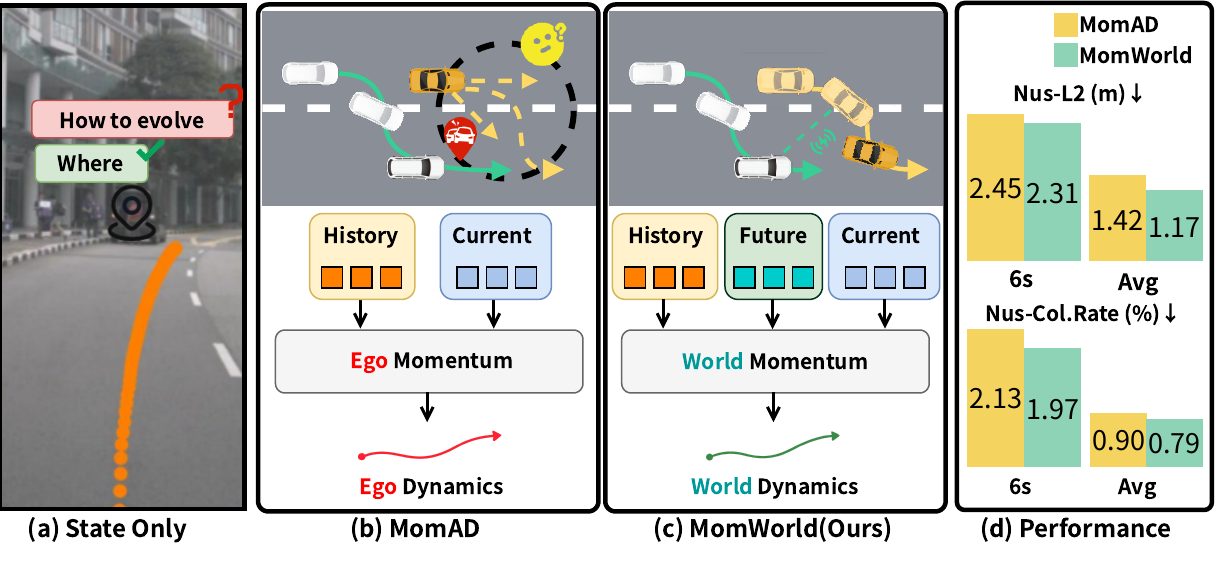}
    \caption{
    \textbf{Motivation and comparison of MomWorld.}
    (a) Reliable future scene evolution remains a key challenge for long-horizon planning.
    (b) MomAD~\citep{song2025momad} derives ego-centric momentum from historical and current evidence to stabilize ego dynamics.
    (c) MomWorld elevates momentum to a world-level latent state, initializes it from past-to-present evidence, and jointly rolls out future configuration and momentum to model world dynamics.
    (d) On six-second nuScenes planning, MomWorld reduces L2 from 2.45 to 2.31\,m and collision rate from 2.13\% to 1.97\%. The corresponding averages improve from 1.42 to 1.17\,m and from 0.90\% to 0.79\%.
    }
    \label{fig:motivation}
\end{figure}

Long-horizon planning has become a central frontier in autonomous driving because
safe decisions depend not only on the current scene, but also on how traffic may
evolve over the next several seconds. Planning-oriented end-to-end systems have
advanced this goal by jointly optimizing perception, prediction, and planning
within a shared representation \citep{hu2023uniad,jiang2023vad,sun2025sparsedrive}.
More recent history-aware methods extend temporal context, while generative
planners enlarge the set of feasible future behaviors
\citep{zhang2025bridgead,song2025momad,liao2025diffusiondrive,song2026diver}.
Despite this progress, increasing the horizon remains difficult because rollout
errors accumulate, interactions such as cut-ins and sudden braking can quickly invalidate
earlier assumptions, and uncertain long-range predictions may perturb otherwise
reliable near-term decisions. Effective long-horizon planning therefore requires
a temporally coherent model of scene evolution rather than a sequence of
independently predicted states.

Latent world models offer a promising route to this objective. By forecasting
compact, task-relevant future representations, they support planning-oriented
imagination without reconstructing every future observation, making multi-step
prediction both efficient and directly useful for decision making
\citep{min2024driveworld,li2025law,zheng2025world4drive,song2026graphworld}.
However, existing planning-oriented latent world models generally do not
explicitly factorize scene configuration and motion trend into jointly propagated
latent states. Repeated transitions may consequently attenuate useful dynamics
during smooth motion, retain stale patterns after abrupt interactions, or
repeatedly infer the same motion cues from scratch. Moreover, conditioning every
planning step on uncertain imagined futures can compromise accurate near-term
decisions. \textbf{The key challenge is therefore to preserve stable trends, refresh them when the scene changes, and introduce future imagination into planning in a controlled manner.}

MomAD~\citep{song2025momad} provides an important insight. Its trajectory and
perception momentum connect historical planning and spatiotemporal evidence to
the current decision, improving temporal continuity between the past and present.
This motivates us to ask whether momentum can connect not only history and the
current scene, but also the predicted future. These momentum mechanisms operate
within perception and planning queries to stabilize the current ego plan. They do
not constitute an explicit latent world state jointly propagated with future
scene configurations. As illustrated in Figure~\ref{fig:motivation}, extending
this idea into future world dynamics can distinguish similar current
configurations that imply different evolutions, while allowing obsolete motion
trends to be suppressed when an interaction changes abruptly.

We introduce \textbf{MomWorld}, a momentum-aware latent world model for
long-horizon planning. MomWorld initializes latent scene state and momentum
from historical-to-current observations, then jointly propagates them across
the predicted future without using future observations at inference. Its first
innovation is a scene-adaptive momentum
rollout. Learnable persistence retains stable trends, scene-conditioned
innovations introduce new dynamics, and a scene-change gate suppresses stale
momentum under abrupt interactions. Its second innovation is \textbf{MoFlow}, a
momentum-conditioned flow-matching module \citep{lipman2023flowmatching} that
refines a strong base trajectory toward futures consistent with the predicted
scene evolution in only a few integration steps. Bounded horizon-aware residual
fusion preserves near-term stability while permitting larger long-range
corrections. Experiments on NAVSIM~\citep{dauner2024navsim,cao2025pseudosimulation}
and nuScenes~\citep{caesar2020nuscenes} demonstrate that MomWorld
improves long-horizon planning consistency and reduces the average predicted
collision rate across the 1--6\,s nuScenes horizon by 12.2\% relative to MomAD.

We make four contributions.
\begin{itemize}
    \item We propose \textbf{MomWorld}, which connects observed history with
    future imagination through Latent World Rollout (LWR).
    \item We introduce \textbf{MoLWM}, which maintains, updates, and resets
    latent momentum to model future dynamics and construct Future World Memory.
    \item We develop \textbf{MoFlow}, which translates predicted world evolution
    into bounded, horizon-aware trajectory corrections for stable planning.
    \item We conduct extensive evaluations on NAVSIM, nuScenes, and Bench2Drive 
    demonstrating stronger long-horizon consistency and a 12.2\% relative
    reduction in average predicted collision rate across the 1--6\,s nuScenes
    horizon over MomAD.
\end{itemize}

\section{Related Work}
\label{sec:related_work}

\subsection{End-to-End Autonomous Driving}

End-to-end autonomous driving jointly learns perception, prediction, and
planning around a shared driving objective. UniAD and VAD establish unified
planning-oriented architectures with query-based and vectorized scene
representations \citep{hu2023uniad,jiang2023vad}. Recent methods improve
computational efficiency through sparse or parallel designs
\citep{sun2025sparsedrive,weng2024paradrive}, exploit historical predictions and
momentum to enhance temporal consistency
\citep{zhang2025bridgead,song2025momad}, and improve multimodal trajectory
generation and selection through truncated diffusion, reinforced diffusion,
constraint-guided flow matching, and generalized trajectory scoring
\citep{liao2025diffusiondrive,song2026diver,liu2025guideflow,li2025generalized}.
These methods primarily improve scene representation, temporal query reuse, or
trajectory generation and selection rather than jointly rolling forward an
explicit latent state of scene configuration and motion dynamics. MomWorld
addresses this gap by rolling past-to-present momentum forward with future
latent world states before trajectory refinement.

\subsection{Latent World Models for Autonomous Driving}

Latent world models encode future scene evolution into compact representations
for planning. DriveWorld learns spatiotemporal features with dynamic memory,
while LAW predicts ego-trajectory-conditioned future features for
self-supervised planning supervision
\citep{min2024driveworld,li2025law}. World4Drive augments physical latents with
spatial-semantic priors and multimodal intentions, whereas Epona jointly models
future video and trajectories through autoregressive diffusion
\citep{zheng2025world4drive,zhang2025epona}. Complementary world-model
directions focus on human and heterogeneous agent dynamics: Driver-WM causally
rolls out in-cabin driver dynamics conditioned on external traffic context
\citep{chi2026driverwm}, whereas PV-WM recurrently co-rolls articulated
pedestrians and rigid vehicles within a synchronized structured state
\citep{pvwm}. Other approaches couple latent imagination with planning more
directly: DriveLaW injects video
latents into a diffusion planner, DriveFuture conditions diffusion planning on
predicted future latents, Latent-WAM autoregressively forecasts compact world
states, and GraphWorld transports an ego-centric relational state before
modulating motion and planning queries
\citep{xia2026drivelaw,hong2026drivefuture,wang2026latentwam,
song2026graphworld}. In contrast, MomWorld jointly rolls out coupled
configuration--momentum states and applies bounded, horizon-aware MoFlow
refinement, preserving near-term stability while enabling stronger long-range
corrections.

\begin{figure*}[t]
\centering
\includegraphics[width=1\textwidth]{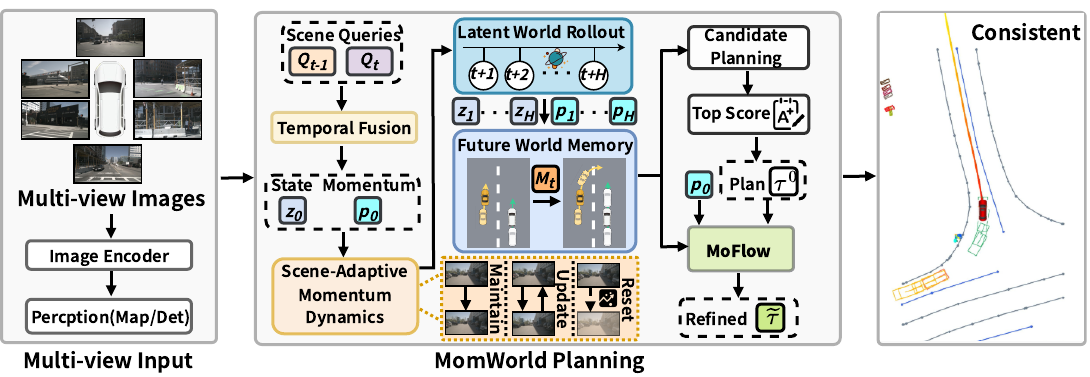}
\caption{Overview of \textbf{MomWorld}. Historical and current multi-view
images are encoded into Scene Queries, from which \textbf{MoLWM} initializes
latent scene state and momentum and propagates them through
Latent World Rollout (LWR) to construct Future World Memory for candidate
scoring and Plan selection. Guided by
history-conditioned momentum and predicted future evolution, \textbf{MoFlow}
applies bounded residual refinement to produce the temporally consistent
Refined trajectory $\widetilde{\tau}$.}
\label{fig:framework}

\end{figure*}

\section{Method}
\label{sec:method}

\subsection{Overview of MomWorld}
\label{sec:method_overview}

MomWorld is an end-to-end autonomous-driving framework designed to improve
long-horizon planning through momentum-aware latent world modeling. As shown in
Fig.~\ref{fig:framework}, historical and current multi-view images are encoded
into Scene Queries, with Map/Det heads retained for perception supervision.
MomWorld comprises two key components. Momentum-Aware Latent World Modeling
uses Latent World Rollout (LWR) to propagate latent scene dynamics and momentum
into Future World Memory for candidate scoring and Plan selection.
Momentum-Conditioned Flow Matching then uses
history-conditioned momentum and predicted future evolution to refine the
selected Plan into a temporally consistent trajectory.

\subsection{Momentum-Aware Latent World Modeling (MoLWM)}
\label{sec:latent_dynamics}
MoLWM represents the latent driving world using a scene state
$z_k\in\mathbb{R}^{D}$ and momentum $p_k\in\mathbb{R}^{D}$, which encode the
scene configuration and its temporal dynamics, respectively. Initialized from
historical and current Scene Queries, MoLWM recursively propagates both
over the planning horizon to construct Future World Memory for Candidate
Planning.
\paragraph{History-Conditioned Initialization.}
Temporal Fusion combines the current scene representation with its
historical variation to initialize $z_0$ and $p_0$.
\begin{equation}
    (z_0,p_0)=
    f_{\mathrm{temp}}\!\left(
    [\operatorname{Pool}(Q_t),
    \operatorname{Pool}(Q_t)-\operatorname{Pool}(Q_{t-1})]
    \right).
\label{eq:world_init}
\end{equation}
The initialized state anchors the current scene configuration, whereas the
initialized momentum captures its history-conditioned evolution.

\paragraph{Scene-Adaptive Momentum Dynamics (SMD).}
To adapt inherited momentum to future scene changes, a multi-head transition
network predicts a retention gate $\rho_k$, a reset gate $g_k$, and a momentum
proposal $u_k$ from the preceding scene state, momentum, and horizon embedding:
\begin{equation}
    (\rho_k,g_k,u_k)
    =
    f_{\mathrm{gate}}
    ([z_{k-1},p_{k-1},r_k^{\mathrm{time}}]).
\label{eq:world_gates}
\end{equation}
The two gates use sigmoid activations, while the momentum proposal uses a
hyperbolic tangent activation. The future momentum is updated by
\begin{equation}
    p_k=
    \operatorname{LN}\!\left(
    \rho_k\odot(1-g_k)\odot p_{k-1}
    +(1-\rho_k)\odot u_k
    \right).
\label{eq:momentum_rollout}
\end{equation}
Here, \emph{Maintain} preserves persistent momentum through $\rho_k$,
\emph{Update} introduces the scene-conditioned proposal $u_k$, and
\emph{Reset} suppresses stale inherited momentum through $g_k$. Together,
these operations produce the scene-adaptive momentum $p_k$ for the $k$-th
future rollout step.

\paragraph{Latent World Rollout (LWR).}
LWR recursively generates paired future State--Momentum predictions from the
history-conditioned pair $(z_0,p_0)$ at the current planning time $t$. At step
$k$, SMD predicts $p_k$ from $(z_{k-1},p_{k-1})$ and the horizon embedding
through Eqs.~\eqref{eq:world_gates}--\eqref{eq:momentum_rollout}. The predicted
momentum then advances the corresponding scene state:
\begin{equation}
    z_k
    =
    z_{k-1}+\Delta t\,\pi_p(p_k),
    \qquad k=1,\ldots,H.
\label{eq:latent_rollout}
\end{equation}
Here, $\pi_p$ maps momentum to a latent transition and $\Delta t$ is the
rollout interval. Feeding each updated pair into the next step produces
$\{(z_k,p_k)\}_{k=1}^{H}$. Taking $t$ as the current time index, the $k$-th
pair corresponds to future step $t+k$ and physical offset $k\Delta t$.
The state $z_k$ predicts the scene configuration, while $p_k$ encodes
history-grounded dynamics adapted to its predicted future evolution.

Algorithm~\ref{alg:future_supervised_rollout} summarizes how training-only
future targets ground the rollout. Future observations are encoded into
detached target Scene Queries:
\begin{equation}
    Q_{t+k}^{\star}
    =
    \operatorname{sg}\!\left(
    \operatorname{ImageEncoder}(\mathcal{O}_{t+k})
    \right),
    \qquad k=1,\ldots,H.
\label{eq:future_targets}
\end{equation}
Here, $\operatorname{sg}(\cdot)$ blocks gradients through the target branch.
The target queries supervise rollout alignment, while future ego, agent, and
presence states provide auxiliary supervision. All future targets are used
only during training and are unavailable at inference.

\begin{wrapfigure}[22]{r}{0.50\textwidth}
\vspace{-0.5pt}
\centering

\begin{minipage}[t]{\linewidth}
\vspace{0pt}

\begin{algorithm}[H]
\caption{Future-Supervised Latent World Rollout}
\label{alg:future_supervised_rollout}
\footnotesize
\DontPrintSemicolon

\KwIn{
Initial state $z_0$ and momentum $p_0$; horizon $H$; future observations
$\{\mathcal{O}_{t+k}\}_{k=1}^{H}$; auxiliary targets
$\{y_{t+k}^{\star}\}_{k=1}^{H}$
}
\KwOut{
Future states and momenta $\{(z_k,p_k)\}_{k=1}^{H}$;
$\mathcal{L}_{\mathrm{future}}$ and
$\mathcal{L}_{\mathrm{aux}}$
}

\textbf{Initialize:}\quad
$\mathcal{L}_{\mathrm{future}},
\mathcal{L}_{\mathrm{aux}}\leftarrow0$\;

\For{$k\leftarrow1$ \KwTo $H$}{

    \textbf{Latent World Rollout:}\quad
    obtain $(z_k,p_k)$ using
    Eqs.~\eqref{eq:world_gates}--\eqref{eq:latent_rollout}\;

    \textbf{Future target:}\quad
    construct $Q_{t+k}^{\star}$ using
    Eq.~\eqref{eq:future_targets}\;

    \textbf{Projection:}\quad
    $\widehat Q_{t+k}\leftarrow f_Q([z_k,p_k])$\;

    \textbf{Alignment:}\quad
    $\mathcal{L}_{\mathrm{future}}\mathrel{+}=
    d_Q(\widehat Q_{t+k},Q_{t+k}^{\star})$\;

    \textbf{Auxiliary loss:}\quad
    $\mathcal{L}_{\mathrm{aux}}\mathrel{+}=
    \ell_{\mathrm{aux}}(z_k,p_k;y_{t+k}^{\star})$\;
}

$\mathcal{L}_{\mathrm{future}}\leftarrow
\mathcal{L}_{\mathrm{future}}/H,
\quad
\mathcal{L}_{\mathrm{aux}}\leftarrow
\mathcal{L}_{\mathrm{aux}}/H$\;

\end{algorithm}

{\footnotesize
\noindent
$f_Q$ projects each LWR step $(z_k,p_k)$ into query space. $d_Q$ aligns future queries,
while $\ell_{\mathrm{aux}}$ supervises future ego, agent, and presence states.
\par}

\end{minipage}
\end{wrapfigure}

\paragraph{Future World Memory (FWM).}
FWM converts each paired LWR prediction into a memory token aligned with its
future step:
\begin{equation}
\begin{gathered}
    m_k=
    \operatorname{LN}\!\left(
    z_k+\operatorname{MLP}([p_k,r_k^{\mathrm{time}}])
    \right),\\[-2pt]
    k=1,\ldots,H.
\end{gathered}
\label{eq:memory_token}
\end{equation}
The residual path preserves the predicted scene state $z_k$, while the learned
branch injects its momentum $p_k$ and temporal position
$r_k^{\mathrm{time}}$. Each token therefore captures the future scene,
dynamics, and horizon information. Dependencies across future steps are
modeled by
\begin{equation}
    M_t=
    \operatorname{SelfAttn}([m_1,\ldots,m_H]).
\label{eq:world_memory}
\end{equation}
Self-attention aggregates context across the rollout, producing a temporally
informed memory at each planning step. During Candidate Planning, each
candidate query cross-attends to $[Q_t;M_t]$. The scoring heads then rank the
decoded candidates and select the Top-Score trajectory as the base Plan
$\tau^0$.

\subsection{Momentum-Conditioned Flow Matching (MoFlow)}
\label{sec:future_planning}
\begin{wrapfigure}{R}{0.5\textwidth}
\vspace{-\intextsep}
\centering

\begin{minipage}[t]{\linewidth}
\vspace{0pt}

{\centering
\setlength{\abovecaptionskip}{4pt}
\setlength{\belowcaptionskip}{0pt}

\includegraphics[
    width=\linewidth
]{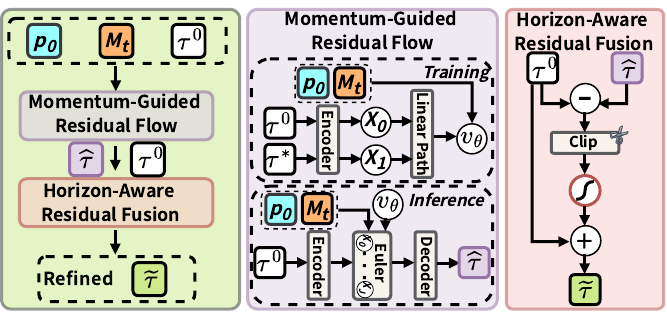}

\caption{
\textbf{Momentum-Conditioned Flow Matching (MoFlow).}
Conditioned on $p_0$ and $M_t$, MoFlow refines the base trajectory through
residual flow and bounded horizon-aware fusion.
}
\label{fig:moflow}
\par}

\end{minipage}
\end{wrapfigure}

MoFlow refines the selected Plan $\tau^0$ using the history-conditioned
momentum $p_0$ and Future World Memory $M_t$. It learns momentum-guided
residuals and applies horizon-aware bounded fusion, as shown in
Fig.~\ref{fig:moflow}, to produce the final trajectory
$\widetilde{\tau}$ while preserving near-term stability.

\paragraph{Momentum-Guided Residual Flow (MGRF).}
The selected Plan $\tau^0$ provides a stable anchor but may not fully capture
the predicted historical-to-future dynamics. MoFlow therefore learns a
residual vector field conditioned on $p_0$ and $M_t$, which provide
history-derived motion and horizon-aligned future dynamics, respectively.

The Trajectory Encoder normalizes positions by a fixed scale
$s_{\mathrm{pos}}=50$ and continuously embeds the heading.
\begin{equation}
    \phi(x,y,\psi)=
    \left(x/s_{\mathrm{pos}},y/s_{\mathrm{pos}},\sin\psi,\cos\psi\right).
\label{eq:trajectory_state}
\end{equation}
Applied waypoint-wise, $\phi$ encodes the base and expert trajectories as
$X_0=\phi(\tau^0)$ and $X_1=\phi(\tau^\star)$, respectively.

During training, we sample $\epsilon\sim\mathcal{U}(0,1)$ and construct the
linear transport path
\begin{equation}
    X_\epsilon=(1-\epsilon)X_0+\epsilon X_1.
\label{eq:flow_path}
\end{equation}
The momentum-conditioned vector field is optimized by
\begin{equation}
    \mathcal{L}_{\mathrm{FM}}
    =
    \mathbb{E}_{\epsilon}
    \left[
    \left\|
    v_\theta(X_\epsilon,\epsilon\mid p_0,M_t)-(X_1-X_0)
    \right\|_2^2
    \right].
\label{eq:flow_loss}
\end{equation}
This formulation learns plan-to-expert residual transport rather than
noise-based trajectory generation.

At inference, we initialize the flow state as $X^0=X_0$ and integrate it
using $J$ explicit Euler steps with midpoint flow-time conditioning.
\begin{equation}
    X^{j+1}
    =
    X^j+\frac{1}{J}
    v_\theta\!\left(X^j,\frac{j+1/2}{J}\mid p_0,M_t\right),
    \qquad j=0,\ldots,J-1.
\label{eq:euler_flow}
\end{equation}
The Trajectory Decoder normalizes the final heading representation and maps it
back to trajectory space.
\begin{equation}
    \widehat{\tau}=\phi^{-1}\!\left(\mathcal{N}_{\psi}(X^J)\right).
\label{eq:flow_output}
\end{equation}

\paragraph{Horizon-Aware Residual Fusion (HARF).}
The flow-refined proposal $\widehat{\tau}$ carries historical-to-future
corrections, but directly replacing $\tau^0$ may compromise near-term
stability. At each horizon step, we first bound its residual.
\begin{equation}
    \Delta\tau_k=
    \operatorname{clip}_{[-\delta,\delta]}
    \left(\widehat{\tau}_k-\tau_k^0\right).
\label{eq:bounded_residual}
\end{equation}
Heading differences are wrapped before clipping. The bounded residual is then
fused with the selected Plan.
\begin{equation}
\begin{aligned}
    \gamma_k
    &=0.05+0.95\frac{k-1}{H-1},\\
    \widetilde{\tau}_k
    &=\tau_k^0+\sigma(b)\gamma_k\Delta\tau_k.
\end{aligned}
\label{eq:residual_fusion}
\end{equation}
Here, $\delta$ bounds each residual component, $b$ is a globally learned
scalar, and $\gamma_k$ linearly increases the correction strength from 0.05
to 1.0 toward longer horizons.

\subsection{Training Objective}
\label{sec:objectives}

Future-world prediction is optimized by
\begin{equation}
    \mathcal{L}_{\mathrm{world}}
    =
    \mathcal{L}_{\mathrm{future}}
    +0.5\mathcal{L}_p
    +0.2\mathcal{L}_{\mathrm{aux}}.
\label{eq:world_loss}
\end{equation}
MoFlow is supervised by
\begin{equation}
    \mathcal{L}_{\mathrm{MoFlow}}
    =
    \mathcal{L}_{\mathrm{FM}}
    +\mathcal{L}_{\tau}.
\label{eq:moflow_loss}
\end{equation}
The complete objective is
\begin{equation}
    \mathcal{L}
    =
    \mathcal{L}_{\mathrm{percep}}
    +\mathcal{L}_{\mathrm{plan}}
    +\mathcal{L}_{\mathrm{world}}
    +\mathcal{L}_{\mathrm{MoFlow}}.
\label{eq:total_loss}
\end{equation}
Here, $\mathcal{L}_{\mathrm{future}}$, $\mathcal{L}_p$, and
$\mathcal{L}_{\mathrm{aux}}$ supervise future queries, momentum, and auxiliary
future states, while $\mathcal{L}_{\mathrm{FM}}$ and $\mathcal{L}_{\tau}$
supervise residual flow and the Refined trajectory.
$\mathcal{L}_{\mathrm{percep}}$ and $\mathcal{L}_{\mathrm{plan}}$ are the
standard Map/Det and candidate-planning losses. Full definitions are provided
in Appendix~1.5.

\section{Experiments}
\label{sec:experiments}


\subsection{Experimental Setup}
\label{sec:exp_setup}

\paragraph{Benchmarks.}
On nuScenes~\citep{caesar2020nuscenes}, we use the official train/validation
split and the camera-only setting. We train separate models for the conventional
3-s horizon (six waypoints) and our 6-s extension (twelve waypoints), both sampled
at 2\,Hz. The long-horizon model is never obtained by extrapolating a 3-s
checkpoint. The 6-s protocol reports planning performance at every second from
1 to 6\,s.
On NAVSIM v1~\citep{dauner2024navsim}, we train on \emph{NavTrain} and evaluate
all 12,146 indexed scenarios of the official \emph{navtest} split with the classic
PDMS. On NAVSIM v2~\citep{cao2025pseudosimulation}, the primary result
is the official two-stage \emph{navhard} pseudo-simulation score with corrected
human-reference filtering. 

Due to space constraints, \textbf{Bench2Drive} results are provided in the Appendix.


\paragraph{Metrics.}
For nuScenes, we report horizon-wise $L_2$ and predicted collision rate, where
lower is better. TPC follows the public MomAD evaluator and measures the
dataset-level Euclidean discrepancy between consecutive trajectory
predictions~\citep{song2025momad}. NAVSIM v1 reports PDMS and its components,
while NAVSIM v2 reports EPDMS and the navhard stage-wise decomposition, where
higher is better.

\paragraph{Implementation Details.}
For NAVSIM, MomWorld is implemented on GTRS-Dense~\citep{li2025generalized}
with a V-99-eSE VoVNet backbone and a $D=256$ latent space. The model takes
two camera frames and four LiDAR sweeps, produces 40 LWR steps at
$\Delta t=0.1$\,s, and scores a fixed vocabulary of 16,384
trajectories. MoFlow refines the selected trajectory using four explicit Euler
steps with midpoint-time conditioning. We train the NAVSIM model for 20 epochs
using Adam with a learning rate of $2\times10^{-4}$, a per-device batch size of
2, mixed precision, and gradient clipping at 5.0. The nuScenes experiments
follow the camera-only 3-s and 6-s settings described above. Controlled
comparisons use identical inputs, backbones, candidate vocabularies, training
data, and optimization budgets within each benchmark.

\subsection{Main Results}
\label{sec:main_results}

\begin{table*}[t]
\scriptsize
\centering
\caption[]{Six-second planning results on the \textbf{nuScenes} validation set. Reported FPS uses an A100 for UniAD, an RTX 3090 for LAW, and an RTX 4090 for SparseDrive, MomAD, and GuideFlow.
}
\renewcommand\arraystretch{0.7}
\setlength{\tabcolsep}{1.99mm}
\resizebox{\linewidth}{!}{%
\begin{tabular}{l c ccccccc ccccccc c}
\toprule
\multirow{2}{*}{$\operatorname{Method}$} &
\multirow{2}{*}{$\operatorname{Venue}$} &
\multicolumn{7}{c}{$\operatorname{L2\ (m)}\downarrow$} &
\multicolumn{7}{c}{$\operatorname{Col.\ Rate\ (\%)}\downarrow$} &
\multirow{2}{*}{$\operatorname{FPS}$} \\
\cmidrule(lr){3-9}\cmidrule(lr){10-16}
&& 1s & 2s & 3s & 4s & 5s & 6s & $\operatorname{Avg.}$
& 1s & 2s & 3s & 4s & 5s & 6s & $\operatorname{Avg.}$ & \\
\midrule
$\operatorname{UniAD}$~\citep{hu2023uniad} & CVPR'23
& 0.47 & 0.91 & 1.35 & 1.91 & 2.47 & 3.07 & \cellcolor{red!5}1.70
& 0.25 & 0.36 & 0.61 & 0.99 & 1.64 & 2.51 & \cellcolor{red!5}1.06 & 1.8 \\
$\operatorname{SparseDrive}$~\citep{sun2025sparsedrive} & ICRA'25
& 0.43 & 0.87 & 1.23 & 1.75 & 2.32 & 2.95 & \cellcolor{red!5}1.59
& 0.19 & 0.31 & 0.56 & 0.87 & 1.54 & 2.33 & \cellcolor{red!5}0.97 & 9.0 \\
$\operatorname{MomAD}$~\citep{song2025momad} & CVPR'25
& 0.41 & 0.85 & 1.13 & 1.67 & 1.98 & 2.45 & \cellcolor{red!5}1.42
& 0.17 & 0.30 & 0.54 & 0.83 & 1.43 & 2.13 & \cellcolor{red!5}0.90 & 7.8 \\
$\operatorname{LAW}$~\citep{li2025law} & ICLR'25
& 0.40 & 0.87 & 1.16 & 1.71 & 2.03 & 2.61 & \cellcolor{red!5}1.46
& 0.19 & 0.33 & 0.57 & 0.86 & 1.51 & 2.31 & \cellcolor{red!5}0.96 & 19.5 \\
$\operatorname{Epona}$~\citep{zhang2025epona} & ICCV'25
& 0.39 & 0.91 & 1.17 & 1.73 & 2.02 & 2.75 & \cellcolor{red!5}1.50
& 0.14 & 0.18 & 0.45 & \textbf{0.74} & 1.48 & 2.23 & \cellcolor{red!5}0.87 & -- \\
$\operatorname{World4Drive}$~\citep{zheng2025world4drive} & ICCV'25
& 0.42 & 0.92 & 1.21 & 1.75 & 2.06 & 2.79 & \cellcolor{red!5}1.53
& 0.16 & 0.20 & 0.47 & 0.76 & 1.50 & 2.14 & \cellcolor{red!5}0.87 & -- \\
$\operatorname{DIVER}$~\citep{song2026diver} & TPAMI'26
& 0.38 & 0.75 & 1.10 & 1.53 & 1.98 & 2.49 & \cellcolor{red!5}1.37
& 0.13 & 0.31 & 0.44 & 0.80 & 1.41 & 2.11 & \cellcolor{red!5}0.87 & 6.6 \\
$\operatorname{GuideFlow}$~\citep{liu2025guideflow} & CVPR'26
& 0.42 & 0.83 & 1.21 & 1.73 & 2.05 & 2.63 & \cellcolor{red!5}1.48
& 0.12 & 0.22 & \textbf{0.42} & 0.79 & 1.44 & 2.15 & \cellcolor{red!5}0.86 & 3.6 \\
\midrule
\cellcolor{red!5}$\operatorname{MomWorld\ (Ours)}$
& \cellcolor{red!5}--
& \cellcolor{red!5}\textbf{0.27} & \cellcolor{red!5}\textbf{0.52}
& \cellcolor{red!5}\textbf{0.86} & \cellcolor{red!5}\textbf{1.28}
& \cellcolor{red!5}\textbf{1.76} & \cellcolor{red!5}\textbf{2.31}
& \cellcolor{red!5}\textbf{1.17}
& \cellcolor{red!5}\textbf{0.02} & \cellcolor{red!5}\textbf{0.17}
& \cellcolor{red!5}\textbf{0.42} & \cellcolor{red!5}0.83
& \cellcolor{red!5}\textbf{1.36} & \cellcolor{red!5}\textbf{1.97}
& \cellcolor{red!5}\textbf{0.79}
& \cellcolor{red!5}7.2 \\
\bottomrule
\end{tabular}}
\label{tab:nusc_6s}
\end{table*}

\begin{table}[H]
\vspace{-0.5\baselineskip}
\noindent
\begin{minipage}[t]{0.46\linewidth}
\vspace{0pt}
\noindent\textbf{Six-second nuScenes planning.}
Table~\ref{tab:nusc_6s} shows that MomWorld achieves the lowest L2 error at all horizons, reducing the best prior average from 1.37 m to 1.17 m. It also obtains the lowest average collision rate of 0.79\%, an 8.1\% improvement over the previous best. Overall, MomWorld provides the best long-horizon planning accuracy and safety.
\end{minipage}\hfill
\begin{minipage}[t]{0.5\linewidth}
\vspace{-0.3cm}
{\centering
\caption[]{Trajectory Prediction Consistency at \textbf{4--6\,s} on the
\textbf{nuScenes} validation set.}
\label{tab:nusc_consistency}
\scriptsize
\renewcommand\arraystretch{0.7}
\setlength{\tabcolsep}{1.2mm}
\resizebox{\linewidth}{!}{%
\begin{tabular}{l c cccc}
\toprule
\multirow{2}{*}{$\operatorname{Method}$} &
\multirow{2}{*}{$\operatorname{Venue}$} &
\multicolumn{4}{c}{$\operatorname{TPC\ (m)}\downarrow$} \\
\cmidrule(lr){3-6}
&& 4s & 5s & 6s & $\operatorname{Avg.}$ \\
\midrule
$\operatorname{UniAD}$~\citep{hu2023uniad} & CVPR'23
& 1.49 & 1.81 & 2.41 & \cellcolor{red!5}1.90 \\
$\operatorname{VAD}$~\citep{jiang2023vad} & ICCV'23
& 1.55 & 1.73 & 2.17 & \cellcolor{red!5}1.82 \\
$\operatorname{SparseDrive}$~\citep{sun2025sparsedrive} & ICRA'25
& 1.33 & 1.66 & 1.99 & \cellcolor{red!5}1.66 \\
$\operatorname{MomAD}$~\citep{song2025momad} & CVPR'25
& 1.19 & 1.45 & 1.61 & \cellcolor{red!5}1.42 \\
\midrule
\cellcolor{red!5}$\operatorname{MomWorld\ (Ours)}$ & \cellcolor{red!5}--
& \cellcolor{red!5}\textbf{0.93} & \cellcolor{red!5}\textbf{1.19}
& \cellcolor{red!5}\textbf{1.46} & \cellcolor{red!5}\textbf{1.19} \\
\bottomrule
\end{tabular}}
\par}
\end{minipage}
\end{table}
\noindent\textbf{Long-Horizon Trajectory Prediction Consistency.}
Table~\ref{tab:nusc_consistency} shows that MomWorld achieves the lowest TPC at all horizons, reducing the best prior average from 1.42 m to 1.19 m. Overall, MomWorld provides the best long-horizon trajectory consistency.

\noindent\textbf{NAVSIM v1 navtest.}
Table~\ref{tab_NAVSIMv1} shows that MomWorld achieves the best PDMS among
world-model methods at 90.2, outperforming DriveLaW by 1.1 points. It also
obtains the highest DAC and EP in this category, demonstrating strong planning
quality under NAVSIM's non-reactive pseudo-simulation protocol.

\noindent\textbf{NAVSIM v2 navtest.}
Table~\ref{tab_navsimv2_navtest} compares MomWorld with TransFuser
\citep{TransFuser}, recent end-to-end planners
\citep{liao2025diffusiondrive,li2025hydraplus,yao2025drivesuprim,zou2025diffusiondrivev2}, VLA methods \citep{liu2026driveworld,li2025recogdrive}, and Latent-WAM \citep{wang2026latentwam}. MomWorld achieves the best EPDMS of 90.1, surpassing Latent-WAM by 0.8 points, while attaining the highest TTC and joint-best EP. These results demonstrate that MomWorld achieves a stronger balance between safety, progress, and overall planning quality
under the diagnostic pseudo-simulation setting.

\noindent\textbf{NAVSIM v2 navhard.}
Table~\ref{tab_navsimv2_navhard} compares MomWorld with end-to-end planners
\citep{TransFuser,liao2025diffusiondrive,liu2025guideflow,feng2026rap,yao2025drivesuprim},
VLA methods \citep{dang2026drivefine,zhou2026spanvla}, and world-model methods
\citep{suna2025minddrive,zheng2025world4drive}. MomWorld achieves the best EPDMS
of 42.8 and the highest DDC in both stages, demonstrating stronger robustness
under perturbed future observations.

\begin{table}[!t]
\vspace{-0.35cm}
\centering
\begin{minipage}[t]{0.49\linewidth}
\vspace{0pt}
\centering
\caption[]{Planning performance on the
\textbf{NAVSIM v1 navtest} split.\ `mmt' denotes the multimodal
TransFuser variant, and $^{*}$ marks our re-implementation.}
\label{tab_NAVSIMv1}
\scriptsize
\renewcommand{\arraystretch}{1.01}
\setlength{\tabcolsep}{0.4mm}
\resizebox{\linewidth}{!}{%
\begin{tabular}{lcccccc}
\toprule
\multicolumn{1}{l}{Method}
& NC$\uparrow$
& DAC$\uparrow$
& TTC$\uparrow$
& Comf.$\uparrow$
& EP$\uparrow$
& PDMS$\uparrow$ \\
\midrule
\rowcolor{cyan!5}
\multicolumn{7}{c}{\textit{E2E-based Methods}} \\
VADv2~\citep{chen2024vadv2}
& 97.2 & 89.1 & 91.6 & \textbf{100} & 76.0 & 80.9 \\
TransFuser~\citep{TransFuser}
& 97.7 & 92.8 & 92.8 & \textbf{100} & 79.2 & 84.0 \\
${\operatorname{TransFuser}_{\operatorname{mmt}}}^{*}$~\citep{TransFuser}
& 96.2 & 95.4 & 90.7 & \textbf{100} & 80.7 & 85.1 \\
UniAD~\citep{hu2023uniad}
& 97.8 & 91.9 & 92.9 & \textbf{100} & 78.8 & 83.4 \\
PARA-Drive~\citep{weng2024paradrive}
& 97.9 & 92.4 & 93.0 & 99.8 & 79.3 & 84.0 \\
DRAMA~\citep{yuan2024drama}
& 98.0 & 93.1 & 94.8 & \textbf{100} & 80.1 & 85.5 \\
Hydra-MDP~\citep{li2024hydra}
& 98.3 & 96.0 & 94.6 & \textbf{100} & 78.7 & 86.5 \\
FUMP~\citep{liu2025fump}
& 98.1 & 96.2 & 94.2 & \textbf{100} & 82.0 & 87.8 \\
DiffusionDrive~\citep{liao2025diffusiondrive}
& 98.2 & 96.2 & 94.7 & \textbf{100} & 82.2 & 88.1 \\
DIVER~\citep{song2026diver}
& \textbf{98.5}
& 96.5
& \textbf{94.9}
& \textbf{100}
& 82.6
& 88.3 \\
DriveSuprim~\citep{yao2025drivesuprim}
& 97.8
& 97.3
& 93.6
& \textbf{100}
& \textbf{86.7}
& 89.9 \\
GoalFlow~\citep{xing2025goalflow}
& 98.4
& \textbf{98.3}
& 94.6
& \textbf{100}
& 85.0
& \textbf{90.3} \\
ReCogDrive-IL~\citep{li2025recogdrive}
& 98.1 & 94.7 & 94.2 & \textbf{100} & 80.9 & 86.5 \\
\midrule
\rowcolor{cyan!5}
\multicolumn{7}{c}{\textit{VLA-based Methods}} \\
AutoVLA~\citep{zhou2025autovla}
& 98.4
& 95.6
& \textbf{98.0}
& 99.9
& 81.9
& 89.1 \\
ReCogdrive~\citep{li2025recogdrive}
& 98.2 & 97.8 & 95.2 & 99.8 & 83.5 & 89.6 \\
DriveWorld-VLA~\citep{liu2026driveworld}
& \textbf{99.1}
& 98.2
& 96.1
& \textbf{100}
& \textbf{85.9}
& \textbf{91.3} \\
\midrule
\rowcolor{cyan!5}
\multicolumn{7}{c}{\textit{World-Model-based Methods}} \\
DrivingGPT~\citep{chen2025drivinggpt}
& 98.9 & 90.7 & 94.9 & 95.6 & 79.7 & 82.4 \\
Resim~\citep{yang2026resim}
& – & – & – & – & – & 86.6 \\
PWM~\citep{georgiev2025pwm}
& 98.6
& 95.9
& 95.4
& \textbf{100}
& 81.8
& 88.1 \\
LAW~\citep{li2025law}
& 96.4 & 95.4 & 88.7 & 99.9 & 81.7 & 84.6 \\
World4Drive~\citep{zheng2025world4drive}
& 97.4
& 94.3
& 92.8
& \textbf{100}
& 79.9
& 85.1 \\
Epona~\citep{zhang2025epona}
& 97.9 & 95.1 & 93.8 & 99.9 & 80.4 & 86.2 \\
WoTE~\citep{wote}
& 98.5 & 96.8 & 94.9 & 99.9 & 81.9 & 88.3 \\
WorldRFT~\citep{yang2026worldrft}
& 97.8
& 96.8
& 94.0
& \textbf{100}
& 81.7
& 87.8 \\
DriveLaW~\citep{xia2026drivelaw}
& \textbf{99.0}
& \textbf{97.1}
& \textbf{96.7}
& \textbf{100}
& 81.3
& 89.1 \\
DriveX-S~\citep{shi2025drivex}
& 97.5
& 94.0
& 93.0
& \textbf{100}
& 79.7
& 84.5 \\
\cellcolor{red!5}
$\operatorname{MomWorld\ (Ours)}$
& \cellcolor{red!5}98.6
& \cellcolor{red!5}\textbf{98.2}
& \cellcolor{red!5}93.9
& \cellcolor{red!5}\textbf{100}
& \cellcolor{red!5}\textbf{85.7}
& \cellcolor{red!5}\textbf{90.2} \\
\bottomrule
\end{tabular}}
\end{minipage}
\hfill
\begin{minipage}[t]{0.49\linewidth}
\vspace{0pt}
\centering
\caption[]{Results on the
\textbf{NAVSIM v2 navtest} split.}
\label{tab_navsimv2_navtest}
\scriptsize
\renewcommand\arraystretch{0.49}
\setlength{\tabcolsep}{0.3mm}
\resizebox{\linewidth}{!}{
\begin{tabular}{l c ccccc cccc}
\toprule
\multicolumn{1}{l}{Method}&  NC$\uparrow$& DAC$\uparrow$& DDC$\uparrow$& TLC$\uparrow$& EP$\uparrow$& TTC$\uparrow$& LK$\uparrow$& HC$\uparrow$& EC$\uparrow$& EPDMS$\uparrow$ \\
        \midrule
        \rowcolor{cyan!5}\multicolumn{11}{c}{\textit{E2E-based Methods}} \\
        TransFuser & 96.9 & 89.9 & 97.8 & 99.7 & 87.1 & 95.4 & 92.7 & 98.3 & 87.2 & 76.7 \\
        DiffusionDrive & 98.2 & 95.9 & 99.4 & 99.8 & 87.5 & 97.3 & 96.8 & 98.3 & 87.7  &84.5 \\
        Hydra-MDP++ & 97.2 & 97.5 & 99.4 & 99.6 & 83.1 & 96.5 & 94.4 & 98.2 & 70.9 & 81.4  \\
        DriveSuprim & 97.5 & 96.5 & 99.4 & 99.6 & 88.4 & 96.6 & 95.5 & 98.3 & 77.0 & 83.1  \\
        DiffusionDriveV2 & 97.7 & 96.6 & 99.2 & 99.8 & 88.9 & 97.2 & 96.0 & 97.8 & 91.0 &  85.5 \\
        \midrule
        \rowcolor{cyan!5}\multicolumn{11}{c}{\textit{VLA-based Methods}} \\
        DriveWorld-VLA & 98.6 & 99.1 & 99.6 & 99.8 & 87.4 & 97.9 & 97.0 & 97.8 & 78.6 & 86.8 \\
        ReCogdrive & 98.3 & 95.2 & 98.3 & 99.8 & 87.1 & 97.5 & 96.6 & 99.5 & 86.5 &  83.6 \\
        \midrule
        \rowcolor{cyan!5}\multicolumn{11}{c}{\textit{World-Model-based Methods}} \\
        Latent-WAM & 98.1 & 97.3 & 99.6 & 99.8 & 87.7 & 97.3 & 97.6 & 98.1 & 87.3 &  89.3 \\
        \rowcolor{red!5}
        MomWorld (Ours)
        & 98.1
        & 98.1
        & 99.5
        & 99.8
        & 88.9
        & 98.2
        & 96.1
        & 98.3
        & 90.6
        & \textbf{90.1} \\
\bottomrule
\end{tabular}}
\par\vspace{0.49\baselineskip}

\caption[]{Results on the \textbf{NAVSIM v2 navhard} split. $^\dagger$ computed from reported metrics.}
\label{tab_navsimv2_navhard}

\scriptsize
\renewcommand{\arraystretch}{0.49}
\setlength{\tabcolsep}{0.25mm}

\resizebox{\linewidth}{!}{%
\begin{tabular}{l c ccccc ccccc}
\toprule
\multicolumn{1}{l}{Method}
& Stage
& NC$\uparrow$
& DAC$\uparrow$
& DDC$\uparrow$
& TLC$\uparrow$
& EP$\uparrow$
& TTC$\uparrow$
& LK$\uparrow$
& HC$\uparrow$
& EC$\uparrow$
& EPDMS$\uparrow$ \\
\midrule

\rowcolor{cyan!5}
\multicolumn{12}{c}{\textit{E2E-based Methods}} \\

\multirow{2}{*}{TransFuser}
& Stage 1
& 96.2 & 79.5 & 99.1 & 99.5 & 84.1
& 95.1 & 94.2 & 97.5 & 79.1
& \multirow{2}{*}{23.1} \\
& Stage 2
& 77.7 & 70.2 & 84.2 & 98.0 & 85.1
& 75.6 & 45.4 & 95.7 & 75.9
& \\

\multirow{2}{*}{DiffusionDrive}
& Stage 1
& 96.0 & 79.7 & 97.4 & 99.5 & 81.3
& 93.1 & 90.8 & 96.8 & 73.8
& \multirow{2}{*}{24.2} \\
& Stage 2
& 82.1 & 72.2 & 88.5 & 98.7 & 85.1
& 78.8 & 49.2 & 89.3 & 71.2
& \\

\multirow{2}{*}{GuideFlow}
& Stage 1
& 96.6 & 80.5 & 96.3 & 99.3 & 82.3
& 94.9 & 91.5 & 97.7 & 67.8
& \multirow{2}{*}{27.1} \\
& Stage 2
& 87.3 & 76.7 & 88.8 & 99.2 & 84.3
& 85.1 & 49.7 & 93.1 & 44.5
& \\

\multirow{2}{*}{RAP-DINO}
& Stage 1
& 97.1 & 94.4 & 98.8 & 99.8 & 83.9
& 96.9 & 94.7 & 96.4 & 66.2
& \multirow{2}{*}{36.9} \\
& Stage 2
& 83.2 & 83.9 & 87.4 & 98.0 & 86.9
& 80.4 & 52.3 & 95.2 & 52.4
& \\

\multirow{2}{*}{DriveSuprim}
& Stage 1
& 98.9 & 95.1 & 99.2 & 99.6 & 76.1
& 99.1 & 94.7 & 97.6 & 54.2
& \multirow{2}{*}{42.1} \\
& Stage 2
& 87.9 & 88.8 & 89.6 & 98.8 & 80.3
& 86.0 & 53.5 & 97.1 & 56.1
& \\

\midrule
\rowcolor{cyan!5}
\multicolumn{12}{c}{\textit{VLA-based Methods}} \\

\multirow{2}{*}{DriveFine}
& Stage 1
& 97.6
& 90.0
& 99.1
& 99.3
& 84.9
& 96.7
& 97.3
& 97.6
& 72.0
& \multirow{2}{*}{30.5$^\dagger$} \\
& Stage 2
& 82.1
& 71.3
& 84.8
& 98.4
& 88.1
& 74.3
& 47.2
& 96.8
& 72.8
& \\

\multirow{2}{*}{SpanVLA}
& Stage 1
& 98.4
& 94.3
& 97.8
& 99.9
& 85.7
& 97.2
& 94.2
& 97.6
& 72.1
& \multirow{2}{*}{40.1} \\
& Stage 2
& 86.9
& 84.3
& 87.1
& 98.2
& 85.5
& 82.7
& 62.3
& 96.8
& 67.4
& \\

\midrule
\rowcolor{cyan!5}
\multicolumn{12}{c}{\textit{World-Model-based Methods}} \\

\multirow{2}{*}{MindDrive}
& Stage 1
& 96.1 & 86.0 & 98.8 & 99.3 & 83.3
& 95.6 & 94.4 & 97.6 & 74.7
& \multirow{2}{*}{30.9} \\
& Stage 2
& 82.6 & 79.1 & 86.4 & 98.0 & 85.3
& 79.4 & 49.2 & 96.5 & 71.0
& \\

\multirow{2}{*}{World4Drive}
& Stage 1
& 97.3 & 89.1 & 97.6 & 99.7 & 60.5
& 96.8 & 87.7 & 93.1 & 60.0
& \multirow{2}{*}{34.9} \\
& Stage 2
& 91.4 & 82.0 & 91.0 & 98.5 & 53.1
& 90.6 & 52.3 & 93.3 & 62.8
& \\

\rowcolor{red!5}
& Stage 1
& 96.9
& 93.6
& 99.8
& 99.8
& 80.4
& 96.9
& 96.4
& 97.6
& 60.0
& \\

\rowcolor{red!5}
\multirow{-2}{*}{\textbf{MomWorld (Ours)}}
& Stage 2
& 86.4
& 88.2
& 94.2
& 98.3
& 81.7
& 84.4
& 55.3
& 97.0
& 54.7
& \multirow{-2}{*}{\textbf{42.8}} \\

\bottomrule
\end{tabular}%
}
\end{minipage}
\end{table}

\subsection{Ablation Studies}
\label{sec:ablations}
\noindent\textbf{Roles of Different Components in MomWorld.}
Table~\ref{tab:component_ablation} shows consistent gains as each component is added. MoLWM progressively improves trajectory accuracy and navhard performance, while MGRF and HARF further reduce long-horizon errors and collisions. The complete MomWorld achieves the best results across all metrics, confirming the complementarity of latent momentum modeling and flow refinement.

\noindent\textbf{Ablation on Different Designs in MoLWM.}
Table~\ref{tab:momentum_mechanism_ablation} shows that removing any MoLWM design degrades performance, with the momentum proposal and reset gate causing the largest drops. The full MoLWM achieves the best results on all metrics, enabling MomWorld to provide more accurate, consistent, and safe planning.

\noindent\textbf{Ablation on Different Designs in MoFlow.}
Table~\ref{tab:moflow_design_ablation} studies the conditioning signals, Euler
steps, residual fusion designs, and inference latency of MoFlow. We first
compare unconditioned refinement with single- and dual-signal conditioning,
showing that jointly using $p_0$ and $M_t$ provides more effective trajectory
guidance. We then vary the number of Euler steps, where four steps achieve the
best accuracy--efficiency trade-off. Finally, removing residual clipping,
horizon-aware weighting, or the learned global gate consistently degrades
performance, while direct trajectory replacement performs worst. Overall, the
full MoFlow achieves the best results with only 8.4\,ms additional latency,
enabling MomWorld to deliver more accurate, safe, and efficient planning.

\begin{table*}[t]
\scriptsize
\centering
\caption[]{\textbf{Roles of Different Method Components in MomWorld.}
Cumulative component study on the \textbf{nuScenes} validation set and the
\textbf{NAVSIM v2 navhard} split.
FWM, LWR, SMD,
MGRF, and HARF denote \emph{Future World Memory},
\emph{Latent World Rollout},
\emph{Scene-Adaptive Momentum Dynamics},
\emph{Momentum-Guided Residual Flow}, and
\emph{Horizon-Aware Residual Fusion}, respectively. LK, EP, and EC are the
Stage-2 metrics reported by the navhard protocol.}
\renewcommand\arraystretch{0.7}
\setlength{\tabcolsep}{1.0mm}
\resizebox{\linewidth}{!}{%
\begin{tabular}{ccccc ccccc cccc}
\toprule
\multicolumn{3}{c}{$\operatorname{MoLWM}$} &
\multicolumn{2}{c}{$\operatorname{MoFlow}$} &
\multicolumn{5}{c}{$\operatorname{nuScenes\ 6s}$} &
\multicolumn{4}{c}{$\operatorname{NAVSIM\ v2\ navhard}$} \\
\cmidrule(lr){1-3}\cmidrule(lr){4-5}\cmidrule(lr){6-10}\cmidrule(lr){11-14}
FWM & LWR & SMD & MGRF & HARF
& \shortstack{$L_2@3\,(\mathrm{m})\downarrow$}
& \shortstack{$L_2@6\,(\mathrm{m})\downarrow$}
& \shortstack{$\operatorname{TPC@6}\,(\mathrm{m})\downarrow$}
& \shortstack{$\operatorname{Col.@3}\,(\%)\downarrow$}
& \shortstack{$\operatorname{Col.@6}\,(\%)\downarrow$}
& \shortstack{$\operatorname{LK}\uparrow$}
& \shortstack{$\operatorname{EP}\uparrow$}
& \shortstack{$\operatorname{EC}\uparrow$}
& \shortstack{$\operatorname{EPDMS}\uparrow$} \\
\midrule
&&&&
& 1.13 & 2.45 & 1.61 & 0.54 & 2.13
& 54.6 & 69.5 & 49.7 & 41.7 \\

\checkmark & & & & 
& 1.07 & 2.42 & 1.57 & 0.51 & 2.09
& 54.8 & 72.3 & 50.8 & 42.0 \\

\checkmark & \checkmark & & & 
& 1.01 & 2.39 & 1.54 & 0.48 & 2.06
& 55.0 & 75.6 & 52.1 & 42.2 \\

\checkmark & \checkmark & \checkmark & & 
& 0.96 & 2.36 & 1.51 & 0.46 & 2.03
& 55.1 & 78.4 & 53.0 & 42.4 \\

\checkmark & \checkmark & \checkmark & \checkmark & 
& 0.91 & 2.33 & 1.48 & 0.44 & 2.00
& 55.2 & 80.2 & 54.0 & 42.6 \\

\rowcolor{red!5}
\checkmark & \checkmark & \checkmark & \checkmark & \checkmark
& \textbf{0.86} & \textbf{2.31} & \textbf{1.46}
& \textbf{0.42} & \textbf{1.97}
& \textbf{55.3} & \textbf{81.7} & \textbf{54.7}
& \textbf{42.8} \\
\bottomrule
\end{tabular}}
\label{tab:component_ablation}
\end{table*}

\begin{table*}[htp]
\centering
\caption[]{Ablation study of different designs in \textbf{MoLWM} across
\textbf{nuScenes}, \textbf{NAVSIM v1 navtest}, and \textbf{NAVSIM v2 navhard}.
}
\label{tab:momentum_mechanism_ablation}
\scriptsize
\renewcommand\arraystretch{0.78}
\setlength{\tabcolsep}{0.55mm}
\resizebox{\linewidth}{!}{%
\begin{tabular}{l ccccc c c}
\toprule
\multirow{2}{*}{$\operatorname{Variant}$}
& \multicolumn{5}{c}{$\operatorname{nuScenes\ 6s}$}
& \multicolumn{1}{c}{$\operatorname{NAVSIM\ v1\ navtest}$}
& \multicolumn{1}{c}{$\operatorname{NAVSIM\ v2\ navhard}$} \\
\cmidrule(lr){2-6}\cmidrule(lr){7-7}\cmidrule(lr){8-8}
& $L_2@3\,(\mathrm{m})\downarrow$
& $L_2@6\,(\mathrm{m})\downarrow$
& $\operatorname{TPC@6}\,(\mathrm{m})\downarrow$
& $\operatorname{Col.@3}\,(\%)\downarrow$
& $\operatorname{Col.@6}\,(\%)\downarrow$
& $\operatorname{PDMS}\uparrow$
& $\operatorname{EPDMS}\uparrow$ \\
\midrule
No historical variation
& 0.95 & 2.58 & 1.72 & 0.55 & 2.40 & 88.6 & 39.7 \\

No horizon embedding in SMD
& 0.90 & 2.43 & 1.57 & 0.47 & 2.12 & 89.5 & 41.3 \\

Fixed retention ($\rho_k\equiv0.9$)
& 0.92 & 2.49 & 1.63 & 0.50 & 2.21 & 89.0 & 40.8 \\

No momentum proposal ($u_k\equiv0$)
& 0.99 & 2.69 & 1.82 & 0.60 & 2.58 & 87.8 & 38.6 \\

No reset gate ($g_k\equiv0$)
& 0.94 & 2.55 & 1.69 & 0.65 & 2.83 & 88.2 & 37.9 \\

No horizon embedding in FWM
& 0.91 & 2.46 & 1.60 & 0.49 & 2.17 & 89.3 & 40.9 \\

No momentum injection in FWM
& 0.97 & 2.64 & 1.78 & 0.58 & 2.49 & 88.0 & 38.9 \\

No temporal aggregation in FWM
& 0.93 & 2.52 & 1.66 & 0.52 & 2.29 & 88.8 & 40.2 \\

\rowcolor{red!5}
\textbf{Full MoLWM}
& \textbf{0.86} & \textbf{2.31} & \textbf{1.46}
& \textbf{0.42} & \textbf{1.97}
& \textbf{90.2} & \textbf{42.8} \\
\bottomrule
\end{tabular}}
\end{table*}

\begin{table*}[t]
\centering
\caption[]{Ablation study of different designs in \textbf{MoFlow} across
\textbf{nuScenes}, \textbf{NAVSIM v1 navtest}, and
\textbf{NAVSIM v2 navhard}.}
\label{tab:moflow_design_ablation}
\scriptsize
\renewcommand\arraystretch{0.78}
\setlength{\tabcolsep}{0.35mm}
\resizebox{\linewidth}{!}{%
\begin{tabular}{l c ccccc c c cc}
\toprule
\multirow{2}{*}{$\operatorname{MoFlow\ Design}$}
& \multirow{2}{*}{$\operatorname{Steps}$}
& \multicolumn{5}{c}{$\operatorname{nuScenes\ 6s}$}
& \multicolumn{1}{c}{$\operatorname{NAVSIM\ v1\ navtest}$}
& \multicolumn{1}{c}{$\operatorname{NAVSIM\ v2\ navhard}$}
& \multicolumn{2}{c}{$\operatorname{Latency\ (ms)}\downarrow$} \\
\cmidrule(lr){3-7}
\cmidrule(lr){8-8}
\cmidrule(lr){9-9}
\cmidrule(lr){10-11}
&
& $L_2@3\,(\mathrm{m})\downarrow$
& $L_2@6\,(\mathrm{m})\downarrow$
& $\operatorname{TPC@6}\,(\mathrm{m})\downarrow$
& $\operatorname{Col.@3}\,(\%)\downarrow$
& $\operatorname{Col.@6}\,(\%)\downarrow$
& $\operatorname{PDMS}\uparrow$
& $\operatorname{EPDMS}\uparrow$
& $\operatorname{Flow}$
& $\operatorname{Total}$ \\
\midrule

No MoFlow (base plan $\tau^0$)
& 0
& 0.96 & 2.36 & 1.51 & 0.46 & 2.03
& 89.7 & 42.4 & 0.0 & 130.5 \\

Unconditioned MGRF
& 4
& 1.01 & 2.53 & 1.67 & 0.57 & 2.38
& 88.3 & 40.0 & 8.4 & 138.9 \\

$M_t$ only
& 4
& 0.91 & 2.40 & 1.55 & 0.47 & 2.09
& 89.4 & 41.8 & 8.4 & 138.9 \\

$p_0$ only
& 4
& 0.93 & 2.43 & 1.58 & 0.49 & 2.14
& 89.2 & 41.5 & 8.4 & 138.9 \\

\midrule

$p_0,M_t$
& 1
& 0.91 & 2.38 & 1.53 & 0.46 & 2.07
& 89.8 & 42.3 & 2.2 & 132.7 \\

$p_0,M_t$
& 2
& 0.88 & 2.34 & 1.49 & 0.44 & 2.01
& 90.0 & 42.5 & 4.3 & 134.8 \\

$p_0,M_t$
& 8
& 0.87 & 2.32 & 1.47 & 0.43 & 1.99
& 90.1 & 42.7 & 16.5 & 147.0 \\

\midrule

Direct replacement ($\widetilde{\tau}=\widehat{\tau}$)
& 4
& 0.98 & 2.57 & 1.69 & 0.59 & 2.46
& 87.9 & 39.5 & 8.4 & 138.9 \\

No residual clipping
& 4
& 0.92 & 2.45 & 1.59 & 0.55 & 2.37
& 88.7 & 40.5 & 8.4 & 138.9 \\

No horizon factor
& 4
& 0.90 & 2.40 & 1.55 & 0.48 & 2.12
& 89.4 & 41.7 & 8.4 & 138.9 \\

No learned global gate ($\sigma(b)\rightarrow1$)
& 4
& 0.89 & 2.38 & 1.53 & 0.47 & 2.09
& 89.6 & 42.0 & 8.4 & 138.9 \\

\rowcolor{red!5}
\textbf{Full MoFlow}
& \textbf{4}
& \textbf{0.86} & \textbf{2.31} & \textbf{1.46}
& \textbf{0.42} & \textbf{1.97}
& \textbf{90.2} & \textbf{42.8}
& 8.4 & 138.9 \\

\bottomrule
\end{tabular}}
\end{table*}

\begin{figure*}[t]
\centering
\includegraphics[
    width=1\textwidth,
]{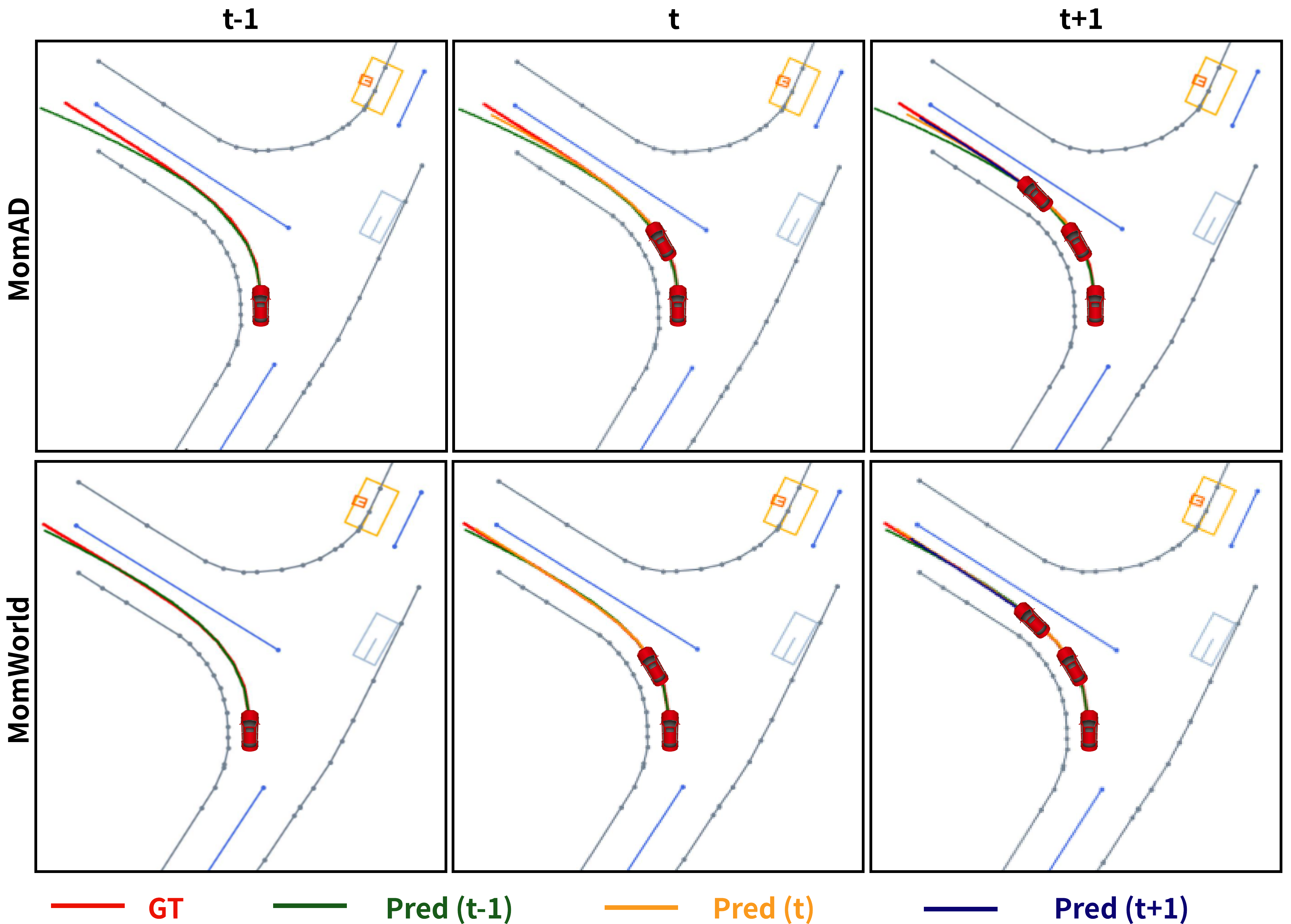}
\caption{
\textbf{Consecutive planning visualization on nuScenes.}
Comparison of MomAD \citep{song2025momad} and MomWorld across consecutive frames.
MomWorld yields more accurate and temporally consistent future trajectories.
}
\label{fig:vis_planning_nus_large}
\end{figure*}




\subsection{Visualization}
\label{sec:Visualization}
Figure~\ref{fig:vis_planning_nus_large} presents consecutive planning results for
MomAD~\citep{song2025momad} and MomWorld. Across $t\!-\!1$, $t$, and $t\!+\!1$, MomWorld exhibits
smoother trajectory evolution and closer alignment with the ground truth,
demonstrating improved temporal stability and planning accuracy.
\section{Conclusion}
\label{sec:conclusion}

We presented \textbf{MomWorld}, a momentum-aware latent world model for
temporally consistent long-horizon autonomous driving. MoLWM jointly propagates
future latent scene states and momentum, using scene-adaptive retention, update,
and reset mechanisms to preserve persistent dynamics while suppressing stale motion.
The resulting Future World Memory supports candidate selection, while MoFlow
refines the selected plan through momentum-conditioned residual flow and
horizon-aware bounded fusion. Experiments on nuScenes, NAVSIM v1/v2, and Bench2Drive  demonstrate improved planning
accuracy, temporal consistency, safety, and planning quality across open-loop
and pseudo-simulation protocols, confirming the complementarity of latent
momentum modeling and controlled flow refinement.

\noindent \textbf{Limitation and Future Work.}
One limitation of MomWorld is its reliance on a fixed candidate vocabulary
and scoring stage. Future work will explore more flexible proposal generation
while retaining stable momentum-conditioned refinement.

\clearpage

\section{Appendix}
\label{app:appendix}

This supplementary material provides additional descriptions and evaluations
of the proposed \textbf{MomWorld} framework. It is organized as follows.
\begin{itemize}
    \item \textbf{Appendix~\ref{app:contributions}} summarizes the
    \textbf{main contributions}.
    \item \textbf{Appendix~\ref{app:broader_impacts}} discusses the
    \textbf{broader impacts}.
    \item \textbf{Appendix~\ref{app:Datasets}} describes the evaluated
    \textbf{datasets and robustness benchmarks}.
    \item \textbf{Appendix~\ref{app:evaluation_metrics}} defines the
    \textbf{evaluation metrics and aggregation protocols}.
    \item \textbf{Appendix~\ref{app:momworld_details}} provides
    \textbf{additional details of MomWorld}.
    \item \textbf{Appendix~\ref{app:implementation}} presents the
    \textbf{implementation and optimization settings}.
    \item \textbf{Appendix~\ref{app:results}} reports
    \textbf{additional planning and robustness results}.
    \item \textbf{Appendix~\ref{app:Qualitative}} provides
    \textbf{additional qualitative planning results}.
\end{itemize}

\subsection{Contributions}
\label{app:contributions}
Our contributions are summarized below.

\noindent\textbf{1) MomWorld Framework.}
We propose \textbf{MomWorld}, a momentum-aware latent world framework for
long-horizon end-to-end autonomous driving. MomWorld explicitly models scene
configuration and momentum, then propagates both from historical-to-current
evidence into the predicted future. This formulation extends momentum from a
cue that stabilizes the current decision into an explicit state for modeling
future world evolution.

\noindent\textbf{2) Momentum-Aware Latent World Modeling.}
We develop \textbf{MoLWM}, which combines History-Conditioned Initialization,
Scene-Adaptive Momentum Dynamics (SMD), and
Latent World Rollout (LWR). Its learned maintain, update, and
reset operations retain persistent trends, introduce scene-conditioned momentum
proposals, and suppress stale inherited momentum when interactions change. The
predicted LWR sequence is organized as Future World Memory (FWM)
for horizon-aligned candidate scoring and base Plan selection.

\noindent\textbf{3) Momentum-Conditioned Flow Matching.}
We introduce \textbf{MoFlow} to translate predicted world evolution into
controlled trajectory corrections. Momentum-Guided Residual Flow (MGRF)
conditions residual transport on history-conditioned momentum and FWM,
enabling efficient refinement with only a few Euler steps.
Horizon-Aware Residual Fusion (HARF) bounds the correction and
gradually increases its influence toward longer horizons, preserving reliable
near-term decisions while allowing stronger long-range refinement.

\noindent\textbf{4) Comprehensive Long-Horizon Evaluation.}
We establish a unified evaluation protocol across six-second nuScenes planning
and NAVSIM pseudo-simulation. The study jointly measures trajectory accuracy,
temporal consistency, predicted collision, planning quality, and computational
cost. Controlled component studies, distribution-shift
benchmarks, and qualitative analysis provide complementary tests of the design
without reducing the evaluation to a single short-horizon score.

\subsection{Broader Impacts}
\label{app:broader_impacts}
MomWorld may improve autonomous-driving safety and reliability by enhancing
the temporal consistency of long-horizon planning. MoLWM preserves stable
motion trends while adapting to scene changes, and MoFlow converts predicted
future evolution into bounded trajectory corrections. These capabilities may
support smoother decisions and safer interactions in complex traffic.
More broadly, momentum-aware latent world modeling may benefit model-based
planning in robotics and embodied intelligence.

\subsection{Datasets}
\label{app:Datasets}

\noindent\textbf{NAVSIM.}
We evaluate MomWorld on the official NAVSIM v1 \textit{navtest} split with
PDMS~\citep{dauner2024navsim} and the official NAVSIM v2 two-stage
\textit{navhard} split with EPDMS~\citep{cao2025pseudosimulation}. For diagnostic
analysis, we additionally apply the one-stage v2 scorer to \textit{navtest}.
This setting is not an official NAVSIM v2 leaderboard protocol. NAVSIM v1 PDMS
combines collision and drivable-area compliance with ego progress,
time-to-collision, and comfort. NAVSIM v2 EPDMS adds driving-direction,
traffic-light, lane-keeping, and extended-comfort criteria. navhard evaluates
an initial real scene together with synthesized follow-up scenes. PDMS and
EPDMS lie in $[0,1]$, and we report them on a 0--100 scale.

\noindent\textbf{Bench2Drive.}
We assess closed-loop performance on Bench2Drive, which contains 220 short
routes spanning 44 interactive scenarios in diverse CARLA environments
\citep{jia2024bench2drive}. We report Driving Score, Success Rate, Efficiency, Comfortness, and the official ability scores for Merging, Overtaking, Emergency Brake, Give Way, and Traffic Sign.

\noindent\textbf{nuScenes.}
We conduct open-loop planning experiments on the official nuScenes validation
split~\citep{caesar2020nuscenes}. The benchmark comprises 1,000 real-world
driving scenes of approximately 20 seconds, captured by a surround-view sensor
suite. Following standard end-to-end planning evaluation and its six-second
extension~\citep{hu2023uniad,song2025momad}, we report L2 displacement error
and predicted collision rate from 1 to 6\,s and TPC over 4--6\,s. Runtime
efficiency is measured in FPS. The independently trained 3-s and 6-s models
predict six and twelve waypoints at 2\,Hz.

\noindent \textbf{Turning-nuScenes (Open-Loop).}
We conduct extensive open-loop experiments on the \textbf{Turning-nuScenes} dataset \citep{song2025momad}, a challenging subset of NuScenes proposed by MomAD \citep{song2025momad} to evaluate trajectory consistency in non-trivial maneuvers. While most planning tasks in the original nuScenes dataset primarily involve go-straight commands, Turning-nuScenes specifically focuses on turning scenarios to assess the temporal coherence of predicted trajectories. To construct this subset, samples are selected using a 25-m displacement threshold between the ground-truth ego positions at 0.5 and 3.0\,s. The resulting validation set comprises 680 samples across 17 scenes, accounting for approximately one-tenth of the full nuScenes validation set.

\noindent \textbf{Adv-nuSc  (Open-Loop).}
To evaluate adversarial robustness, we conduct extensive open-loop experiments on the \textbf{Adv-nuSc} \citep{xu2025challenger} dataset. It contains 156 scenes (6,115 samples) and is specifically crafted to challenge the ego vehicle by introducing adversarial traffic participants. It is built upon the validation split of the nuScenes dataset  \citep{caesar2020nuscenes}, which contains 150 scenes, each with 20 seconds of driving data. For each scene, we randomly select up to 10 background vehicles (if there are that many) that come close to the ego vehicle at any point in time and designate them as candidate adversarial agents. Challenger is then used to generate adversarial trajectories for these vehicles, creating diverse and challenging driving scenarios.

\noindent \textbf{NuScenes-C  (Open-Loop).}
\textbf{NuScenes-C} \citep{dong2023corruptions} is a corrupted benchmark derived from the nuScenes validation set, introducing various types of noise to assess the robustness of planning models. It includes 27 corruption types applied at 5 severity levels. To evaluate robustness under adverse weather conditions, we select three representative weather corruptions — Rain, Snow, and Fog — as our test scenarios.

\subsection{Evaluation Metrics}
\label{app:evaluation_metrics}

\noindent\textbf{nuScenes, Turning-nuScenes, Adv-nuSc, and nuScenes-C.}
Let $\widehat{\mathbf y}_{n,T}$ and $\mathbf y_{n,T}$ denote the predicted and
expert ego positions of sample $n$ at horizon $T$. The horizon-wise displacement
error is
\begin{equation}
    \mathrm{L2}@T=
    \frac{1}{N}\sum_{n=1}^{N}
    \left\|
    \widehat{\mathbf y}_{n,T}-\mathbf y_{n,T}
    \right\|_2 .
\label{eq:app_nusc_l2}
\end{equation}
The average L2 is the arithmetic mean over the reported horizons.

For collision evaluation, let $\widehat{\mathcal B}_{n,h}$ denote the oriented
ego box induced by the predicted trajectory and $\mathcal B^{a}_{n,h}$ the box
of participant $a$. Over the valid sample set $\mathcal V_T$, the predicted
collision rate is
\begin{equation}
    \mathrm{Col.}@T=
    \frac{100}{|\mathcal V_T|}
    \sum_{n\in\mathcal V_T}
    \mathbb{I}\!\left[
    \exists\,h\leq T,\ a:
    \widehat{\mathcal B}_{n,h}\cap
    \mathcal B^{a}_{n,h}\neq\varnothing
    \right].
\label{eq:app_nusc_collision}
\end{equation}
Samples whose expert trajectories are already in collision are excluded.

Trajectory Prediction Consistency (TPC) follows the public MomAD
protocol~\citep{song2025momad}. Let
$\widehat{\mathbf y}_{n,k}^{\,t}$ and
$\widehat{\mathbf y}_{n,k}^{\,t-1}$ denote the $k$-th waypoint from two
consecutive planning outputs under the evaluator coordinate convention.
For the $K_T$ waypoints up to horizon $T$, TPC is
\begin{equation}
    \mathrm{TPC}@T=
    \frac{1}{N K_T}
    \sum_{n=1}^{N}\sum_{k=1}^{K_T}
    \left\|
    \widehat{\mathbf y}_{n,k}^{\,t}
    -
    \widehat{\mathbf y}_{n,k}^{\,t-1}
    \right\|_2.
\label{eq:app_tpc}
\end{equation}
Lower L2, Pred.-Col., and TPC indicate greater accuracy, safety, and temporal
stability.

\noindent\textbf{NAVSIM v1.}
NAVSIM v1 combines no-at-fault collision (NC), drivable-area compliance (DAC),
ego progress (EP), time-to-collision (TTC), and comfort (C) into PDMS:
\begin{equation}
    \mathrm{PDMS}
    =
    \mathrm{NC}\,\mathrm{DAC}
    \frac{
    5\,\mathrm{EP}
    +5\,\mathrm{TTC}
    +2\,\mathrm{C}
    }{12}.
\label{eq:app_pdms}
\end{equation}
All components lie in $[0,1]$, while the tables report percentage-scaled
values. Higher PDMS indicates better overall planning quality.

\noindent\textbf{NAVSIM v2.}
NAVSIM v2 introduces driving-direction compliance (DDC), traffic-light
compliance (TLC), lane keeping (LK), history comfort (HC), and extended comfort
(EC). To avoid penalizing behavior also exhibited by the human reference, each
component is filtered as
\begin{equation}
    f_m=
    \begin{cases}
        1, & m_{\mathrm{human}}=0,\\
        m_{\mathrm{agent}}, & \text{otherwise}.
    \end{cases}
\label{eq:app_human_filter}
\end{equation}
The single-stage EPDMS is
\begin{equation}
    \mathrm{EPDMS}
    =
    f_{\mathrm{NC}}f_{\mathrm{DAC}}
    f_{\mathrm{DDC}}f_{\mathrm{TLC}}
    \frac{
    5f_{\mathrm{EP}}+5f_{\mathrm{TTC}}
    +2f_{\mathrm{LK}}+2f_{\mathrm{HC}}+2f_{\mathrm{EC}}
    }{16}.
\label{eq:app_epdms}
\end{equation}

For navhard, let $s_1$ be the first-stage EPDMS and $s_{2,i}$ the score of
follow-up scene $i$. Its relevance to the first-stage endpoint
$\widehat{\mathbf x}$ is determined by
\begin{equation}
    \bar w_i=
    \frac{
    \exp\!\left(
    -\|\mathbf x_i-\widehat{\mathbf x}\|_2^2/(2\sigma^2)
    \right)}
    {
    \sum_j
    \exp\!\left(
    -\|\mathbf x_j-\widehat{\mathbf x}\|_2^2/(2\sigma^2)
    \right)} .
\label{eq:app_navhard_weight}
\end{equation}
The second-stage score is the Gaussian-weighted aggregation
\begin{equation}
    s_2=\sum_i \bar w_i s_{2,i}.
\label{eq:app_navhard_stage2}
\end{equation}
The final navhard score multiplies the two stages:
\begin{equation}
    \mathrm{EPDMS}_{\mathrm{navhard}}=s_1s_2.
\label{eq:app_navhard_final}
\end{equation}
Thus, navhard EPDMS is not an arithmetic mean of the stage-wise scores.

\noindent\textbf{Bench2Drive.}
Let $N_{\mathrm{route}}$ be the number of evaluated routes and
$N_{\mathrm{succ}}$ the number completed without infractions. Success Rate is
\begin{equation}
    \mathrm{SR}
    =
    \frac{N_{\mathrm{succ}}}{N_{\mathrm{route}}}\times100.
\label{eq:app_b2d_sr}
\end{equation}
Driving Score combines the route-completion percentage
$RC_i\in[0,100]$ with the multiplicative infraction penalties $p_{i,j}$:
\begin{equation}
    \mathrm{DS}
    =
    \frac{1}{N_{\mathrm{route}}}
    \sum_{i=1}^{N_{\mathrm{route}}}
    \mathrm{RC}_i
    \prod_{j=1}^{K_i}p_{i,j}.
\label{eq:app_b2d_ds}
\end{equation}

Efficiency compares ego speed with the mean speed of nearby traffic at the
official route checkpoints:
\begin{equation}
    \mathrm{Effi.}
    =
    \frac{100}{Q}
    \sum_{q=1}^{Q}
    \frac{v_q^{\mathrm{ego}}}{v_q^{\mathrm{near}}}.
\label{eq:app_b2d_efficiency}
\end{equation}
Comfortness evaluates whether acceleration, yaw, and jerk variables remain
within their prescribed bounds. With $\mathcal S$ denoting the evaluated
segments and $\mathcal R$ the set of smoothness variables, it is summarized as
\begin{equation}
    \mathrm{Comf.}
    =
    \frac{100}{|\mathcal S|}
    \sum_{s\in\mathcal S}
    \prod_{t\in s}\prod_{r\in\mathcal R}
    \mathbb{I}
    \!\left[l_r\leq q_{t,r}\leq u_r\right].
\label{eq:app_b2d_comfort}
\end{equation}

For each driving ability $a$, the ability score is the success rate over its
corresponding route subset $\mathcal R_a$:
\begin{equation}
    \mathrm{Ability}_a
    =
    \frac{100}{|\mathcal R_a|}
    \sum_{i\in\mathcal R_a}
    \mathbb{I}[\mathrm{success}_i].
\label{eq:app_b2d_ability}
\end{equation}
We report Merging, Overtaking, Emergency Brake, Give Way, and Traffic Sign.
Higher values are better for all Bench2Drive metrics.

\subsection{Additional Details of MomWorld}
\label{app:momworld_details}

\noindent\textbf{Latent World Rollout (LWR).}
LWR starts from the history-conditioned scene state $z_0$ and momentum $p_0$,
which are initialized from the current Scene Query and its historical variation.
\begin{equation}
    (z_0,p_0)=
    f_{\mathrm{temp}}\!\left(
    [\operatorname{Pool}(Q_t),
    \operatorname{Pool}(Q_t)-\operatorname{Pool}(Q_{t-1})]
    \right).
\label{eq:app_world_init}
\end{equation}
At each future step, Scene-Adaptive Momentum Dynamics uses the preceding
rollout variables and the horizon embedding to produce feature-wise retention,
reset, and innovation signals.
Writing $x_k=[z_{k-1},p_{k-1},r_k^{\mathrm{time}}]$ and denoting its three
heads by $f_{\rho}$, $f_g$, and $f_u$, the complete transition is
\begin{equation}
\begin{aligned}
    \rho_k&=\sigma(f_{\rho}(x_k)),
    &g_k&=\sigma(f_g(x_k)),
    &u_k&=\tanh(f_u(x_k)),\\
    p_k&=\operatorname{LN}\!\left(
    \rho_k\odot(1-g_k)\odot p_{k-1}
    +(1-\rho_k)\odot u_k
    \right),\\
    z_k&=z_{k-1}+\Delta t\,\pi_p(p_k),
    &k&=1,\ldots,H.
\end{aligned}
\label{eq:app_lwr_transition}
\end{equation}
The inherited branch $\rho_k\odot(1-g_k)\odot p_{k-1}$ preserves persistent
motion while suppressing stale components after a scene change. The innovation
branch $(1-\rho_k)\odot u_k$ introduces dynamics supported by the current
rollout context. Layer normalization controls the momentum scale during long
unrolling, and $\pi_p$ converts each momentum update into a latent scene-state
increment. Repeated application of Eq.~\eqref{eq:app_lwr_transition} produces
the complete future rollout.

The LWR output is the horizon-aligned sequence
$\{(z_k,p_k)\}_{k=1}^{H}$. As detailed in
Algorithm~\ref{alg:fwm_candidate_scoring}, each step is converted into a memory
token before temporal aggregation organizes the sequence into Future World
Memory $M_t$. At inference, the entire sequence is generated from historical
and current Scene Queries without access to future observations.

\noindent\textbf{Temporal Validity Handling.}
A deterministic indicator $v_t\in\{0,1\}$ specifies whether the preceding
Scene Query is temporally valid. The historical input to Temporal Fusion is
defined as
\begin{equation}
    Q_{t-1}^{\mathrm{in}}
    =
    v_tQ_{t-1}+(1-v_t)Q_t.
\label{eq:app_temporal_fallback}
\end{equation}
We set $v_t=0$ at scene boundaries, for missing predecessors, or when the
expected frame interval is violated. In these cases,
$Q_{t-1}^{\mathrm{in}}=Q_t$, yielding zero temporal variation and preventing
invalid history from affecting $(z_0,p_0)$. Latent state, momentum, and Future
World Memory are not reused across scene boundaries.

\noindent\textbf{Future World Memory.}
Future World Memory is reconstructed from the predicted LWR sequence at each
planning step rather than maintained across frames.
Algorithm~\ref{alg:fwm_candidate_scoring} details how this memory conditions
candidate representations for scoring and selection of the base Plan
$\tau^0$.

\noindent\textbf{Future-Target Supervision.}
Future observations are encoded by the shared Image Encoder and detached to
construct target Scene Queries:
\begin{equation}
    Q_t^\star=\operatorname{sg}(Q_t),
    \qquad
    Q_{t+k}^{\star}
    =
    \operatorname{sg}\!\left(
    \operatorname{ImageEncoder}(\mathcal{O}_{t+k})
    \right),
    \quad k=1,\ldots,H,
\label{eq:app_future_targets}
\end{equation}
where $\operatorname{sg}(\cdot)$ blocks gradient propagation through the
target-query branch. The latent rollout is aligned with these targets through
\begin{equation}
    \mathcal{L}_{\mathrm{future}}
    =
    \frac{1}{H}\sum_{k=1}^{H}
    d_Q\!\left(
    f_Q([z_k,p_k]),Q_{t+k}^{\star}
    \right),
\label{eq:app_future_loss}
\end{equation}
where $f_Q$ projects each LWR step $(z_k,p_k)$ into the Scene Query
space and $d_Q$ denotes the matching loss. Future momentum is supervised by
consecutive target-query transitions:
\begin{equation}
    \mathcal{L}_p
    =
    \frac{1}{H}\sum_{k=1}^{H}
    \left\|
    f_\Delta(p_k)
    -
    \left[
    \operatorname{Pool}(Q_{t+k}^{\star})
    -
    \operatorname{Pool}(Q_{t+k-1}^{\star})
    \right]
    \right\|_1,
\label{eq:app_momentum_loss}
\end{equation}
where $f_\Delta$ projects latent momentum into the pooled-query space.
Auxiliary future-state supervision is defined as
\begin{equation}
    \mathcal{L}_{\mathrm{aux}}
    =
    \frac{1}{H}\sum_{k=1}^{H}
    \left(
    \ell_{\mathrm{ego}}^k
    +\ell_{\mathrm{agent}}^k
    +\ell_{\mathrm{pres}}^k
    \right).
\label{eq:app_auxiliary_loss}
\end{equation}
These terms supervise future ego states, agent states, and participant
presence. All future observations and annotations are used only during
training and are unavailable at inference.

\noindent\textbf{MoFlow Trajectory Supervision.}
Trajectory waypoints are encoded using normalized positions and a continuous
heading representation:
\begin{equation}
    \phi(x,y,\psi)=
    \left(x/s_{\mathrm{pos}},y/s_{\mathrm{pos}},\sin\psi,\cos\psi\right).
\label{eq:app_trajectory_state}
\end{equation}
Let $X_0=\phi(\tau^0)$ and $X_1=\phi(\tau^\star)$ denote the encoded base and
expert trajectories. For $\epsilon\sim\mathcal{U}(0,1)$, we form
$X_\epsilon=(1-\epsilon)X_0+\epsilon X_1$ and optimize
\begin{equation}
    \mathcal{L}_{\mathrm{FM}}
    =
    \mathbb{E}_{\epsilon}\!\left[
    \left\|
    v_\theta(X_\epsilon,\epsilon\mid p_0,M_t)-(X_1-X_0)
    \right\|_2^2
    \right].
\label{eq:app_flow_loss}
\end{equation}
This loss supervises the momentum-conditioned residual vector field. The final
Refined trajectory is supervised by
\begin{equation}
    \mathcal{L}_{\tau}
    =
    \frac{1}{H}\sum_{k=1}^{H}
    \left\|
    \widetilde{\tau}_k-\tau_k^\star
    \right\|_1,
\label{eq:app_trajectory_loss}
\end{equation}
where $\tau^\star$ denotes the expert trajectory. The standard perception and
candidate-planning objectives retain the definitions of the underlying
planner.
The complete training objective is
\begin{equation}
\begin{aligned}
\mathcal{L}_{\mathrm{world}}
&=
\mathcal{L}_{\mathrm{future}}
+0.5\mathcal{L}_{p}
+0.2\mathcal{L}_{\mathrm{aux}},\\
\mathcal{L}_{\mathrm{MoFlow}}
&=
\mathcal{L}_{\mathrm{FM}}
+\mathcal{L}_{\tau},\\
\mathcal{L}
&=
\mathcal{L}_{\mathrm{percep}}
+\mathcal{L}_{\mathrm{plan}}
+\mathcal{L}_{\mathrm{world}}
+\mathcal{L}_{\mathrm{MoFlow}}.
\end{aligned}
\label{eq:app_total_loss}
\end{equation}

\begin{algorithm}[H]
\caption{Future-Memory-Conditioned Candidate Scoring}
\label{alg:fwm_candidate_scoring}
\footnotesize
\DontPrintSemicolon

\KwIn{
Current Scene Queries $Q_t$;
predicted future pairs $\{(z_k,p_k)\}_{k=1}^{H}$;
candidate vocabulary $\mathcal{V}=\{\tau_i\}_{i=1}^{N}$;
ego status $s_t$
}
\KwOut{
Future World Memory $M_t$;
candidate scores $\{S_i\}_{i=1}^{N}$;
base Plan $\tau^0$
}

\For{$k\leftarrow1$ \KwTo $H$}{
    \textbf{Memory token:}\quad
    $m_k\leftarrow
    \operatorname{LN}\!\left(
    z_k+\operatorname{MLP}([p_k,r_k^{\mathrm{time}}])
    \right)$\;
}

\textbf{Temporal memory:}\quad
$M_t\leftarrow
\operatorname{SelfAttn}([m_1,\ldots,m_H])$\;

\textbf{Scoring context:}\quad
$C_t\leftarrow\operatorname{Concat}(Q_t,M_t)$\;

\For{$i\leftarrow1$ \KwTo $N$}{
    \textbf{Candidate encoding:}\quad
    $e_i\leftarrow
    f_\tau(\operatorname{vec}(\tau_i))$\;
}

\textbf{Candidate interaction:}\quad
$E_\tau\leftarrow
\operatorname{CandEnc}([e_1,\ldots,e_N])$\;

\textbf{Future-conditioned decoding:}\quad
$H_t^\tau\leftarrow
\operatorname{TrajDecoder}(E_\tau,C_t)$\;

\For{$i\leftarrow1$ \KwTo $N$}{
    \textbf{Status conditioning:}\quad
    $h_{t,i}\leftarrow H_{t,i}^\tau+W_s s_t$\;

    \textbf{Scoring heads:}\quad
    $\ell_i^r\leftarrow g_r(h_{t,i}),
    \quad r\in\mathcal{R}$\;
}

$\mathcal{R}\leftarrow
\{\mathrm{imi},\mathrm{NC},\mathrm{DAC},\mathrm{TTC},
\mathrm{EP},\mathrm{DDC},\mathrm{LK},\mathrm{TLC}\}$\;

\textbf{Imitation distribution:}\quad
$\pi^{\mathrm{imi}}\leftarrow
\operatorname{softmax}(\ell^{\mathrm{imi}})$\;

\For{$i\leftarrow1$ \KwTo $N$}{
    \textbf{Rule probabilities:}\quad
    $p_i^r\leftarrow\sigma(\ell_i^r),
    \quad r\in\mathcal{R}\setminus\{\mathrm{imi}\}$\;

    \textbf{Score aggregation:}\quad
    $S_i\leftarrow
    \operatorname{Aggregate}
    (\pi_i^{\mathrm{imi}},\{p_i^r\})$\;
}

\textbf{Plan selection:}\quad
$i^\star\leftarrow\arg\max_i S_i$,
$\quad\tau^0\leftarrow\tau_{i^\star}$\;

\end{algorithm}

Each candidate trajectory is represented by a query token, while the
concatenated current-scene and future-memory tokens $C_t=[Q_t;M_t]$ serve as
the key--value context of the trajectory decoder. For attention head $a$, this
interaction is written as
\begin{equation}
\begin{aligned}
    A_t^{(a)}
    &=
    \operatorname{softmax}\!\left(
    \frac{
    (E_\tau W_Q^{(a)})
    (C_tW_K^{(a)})^\top
    }{\sqrt{d_a}}
    \right),\\
    Z_t^{(a)}
    &=
    A_t^{(a)}(C_tW_V^{(a)}).
\end{aligned}
\label{eq:app_fwm_cross_attention}
\end{equation}
Although all candidates share the same $M_t$, their distinct query tokens
produce different attention weights. The resulting representations therefore
capture candidate-specific compatibility with the predicted scene evolution.

In our NAVSIM implementation, $N=16{,}384$, $H=40$, and $D=256$. Each
candidate contains $40$ poses over a $4$-s horizon, while $M_t$ contains one
future-memory token per planning step. The trajectory decoder contains three
layers with eight attention heads. During training, vocabulary dropout retains
half of the candidates, whereas all candidates are scored at inference.

The predicted logits correspond to imitation, no-at-fault collision (NC),
drivable-area compliance (DAC), time-to-collision (TTC), ego progress (EP),
driving-direction compliance (DDC), lane keeping (LK), and traffic-light
compliance (TLC). Their inference score is
\begin{equation}
\begin{aligned}
    S_i
    &=0.03\log\pi_i^{\mathrm{imi}}
    +0.10\log p_i^{\mathrm{TLC}}
    +0.10\log p_i^{\mathrm{NC}}
    +0.90\log p_i^{\mathrm{DAC}}\\
    &\quad
    +0.20\log p_i^{\mathrm{DDC}}
    +6.0\log\!\left(
        7.0p_i^{\mathrm{TTC}}
        +7.0p_i^{\mathrm{EP}}
        +3.0p_i^{\mathrm{LK}}
    \right).
\end{aligned}
\label{eq:app_candidate_score}
\end{equation}
The scoring heads are supervised by candidate-level imitation targets and PDM
subscores, allowing gradients to propagate through the trajectory decoder into
$M_t$ and thereby learn future-conditioned candidate ranking.

Candidate scoring is performed only once at each planning step. Each complete
candidate is represented by one query token that attends globally to the
future-memory sequence; no candidate-specific world rollout or explicit
one-to-one temporal matching is performed. After selecting $\tau^0$, MoFlow
uses the same $M_t$ to refine this Plan without reconstructing the memory or
re-ranking the candidate vocabulary.

\subsection{Implementation Details}
\label{app:implementation}

\begin{table}[t]
\centering
\scriptsize
\caption{NAVSIM implementation configuration of MomWorld.}
\label{tab:app_config}
\renewcommand{\arraystretch}{0.85}
\setlength{\tabcolsep}{3.0pt}
\begin{tabular*}{\linewidth}{@{\extracolsep{\fill}}ll@{}}
\toprule
Configuration & Value \\
\midrule
Model input history & 2 camera frames / 4 LiDAR sweeps \\
Image resolution & $2048\times512$ \\
Image backbone & V-99-eSE VoVNet \\
Planner latent / FFN width & 256 / 1024 \\
Planner attention layers / heads & 3 / 8 \\
Candidate vocabulary size & 16,384 \\
Planning horizon / interval & 40 steps / 0.1\,s \\
LWR horizon / interval & 40 steps / 0.1\,s \\
Maximum agent slots & 30 \\
Momentum persistence initialization & 0.9 \\
Trajectory encoding
& Eq.~\eqref{eq:app_trajectory_state}, $s_{\mathrm{pos}}=50$ \\
Flow solver / steps & explicit Euler with midpoint-time conditioning / 4 \\
Heading normalization & after Euler integration \\
Residual clamp $\delta$ & componentwise, $\pm2.0$ \\
Residual-gate initialization $b$ & $-4$ \\
Horizon weighting & linear ramp, $0.05\!\rightarrow\!1.0$ \\
Optimizer / learning rate & Adam / $2\times10^{-4}$ \\
Epochs / per-device batch size & 20 / 2 \\
Precision / gradient clipping & FP16 mixed precision / 5.0 \\
\bottomrule
\end{tabular*}
\end{table}

\noindent\textbf{NAVSIM Instantiation.}
Table~\ref{tab:app_config} summarizes the complete NAVSIM configuration.
We instantiate MomWorld on GTRS-Dense~\citep{li2025generalized} and keep the
sensor input, backbone, candidate vocabulary, and 4-s planning grid fixed
across all controlled comparisons.

\noindent\textbf{MoLWM Configuration.}
MoLWM uses LWR to propagate latent scene states and momentum on the planner
grid.
Future ego trajectories and track-aligned agent state and presence annotations
are used only as training targets and are unavailable at inference.

\noindent\textbf{MoFlow Configuration.}
MoFlow refines the selected Plan using four explicit Euler updates with
midpoint flow-time conditioning. After the four Euler updates, the heading
representation is normalized once before trajectory decoding. HARF applies a
componentwise residual clamp followed by a linear horizon ramp, while the
negative gate initialization starts training from conservative corrections.

\noindent\textbf{Optimization Protocol.}
The perception, planning, MoLWM, and MoFlow objectives are jointly optimized
with Eq.~\eqref{eq:app_total_loss}. Unless otherwise specified, controlled
ablations use identical inputs, backbones, candidate trajectories, training
data, and optimization budgets.

\begin{figure*}[t]
  \centering
  \includegraphics[width=\linewidth]{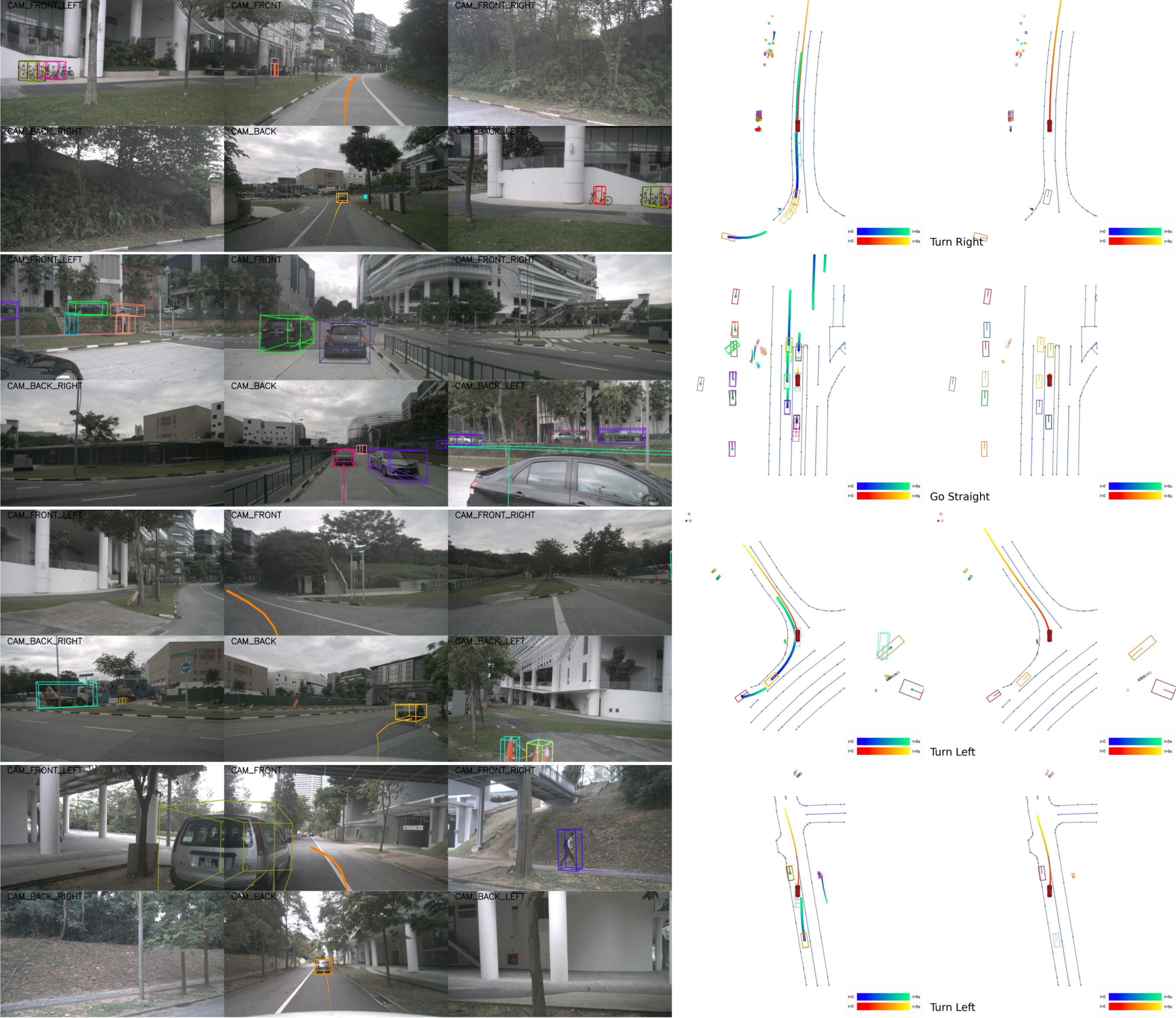}
  \caption{
  \textbf{Qualitative 6-second planning results on nuScenes.}
  MomWorld generates smooth trajectories when transitioning from turns to
  straight driving, decelerating in dense traffic, turning left at
  intersections, and avoiding vehicles ahead.
  }
  \label{nus_vis_fl}
\end{figure*}

\subsection{More Planning Results}
\label{app:results}

\begin{table*}[t]
\centering
\caption{Closed-loop planning and multi-ability performance on
\textbf{Bench2Drive}. $^{*}$ denotes expert feature distillation. Higher is
better for all metrics.}
\label{tab_b2d}
\scriptsize
\renewcommand{\arraystretch}{0.82}
\setlength{\tabcolsep}{0.65mm}
\resizebox{\linewidth}{!}{%
\begin{tabular}{l l cccc cccccc}
\toprule
\multirow{2}{*}{Method}
& \multirow{2}{*}{Venue}
& \multicolumn{4}{c}{Closed-loop Performance}
& \multicolumn{6}{c}{Multi-Ability (\%)$\uparrow$} \\
\cmidrule(lr){3-6}\cmidrule(lr){7-12}
& & DS$\uparrow$ & SR (\%)$\uparrow$ & Effi.$\uparrow$ & Comf.$\uparrow$
& Merge & Overtake & \shortstack{Emergency Brake} & \shortstack{Give Way}
& \shortstack{Traffic Sign} & Mean \\
\midrule

TCP-traj$^{*}$~\citep{wu2022tcp} & NeurIPS'22 & 59.90 & 30.00 & 76.54  & 18.08 & 12.50 & 22.73 & 52.72 & 40.00 & 46.63 & 34.92 \\
UniAD~\citep{hu2023uniad} & CVPR'23 & 45.81 & 16.36 & 129.21 & 43.58 & 14.10 & 17.78 & 21.67 & 10.00 & 14.21 & 15.55 \\
ThinkTwice$^{*}$~\citep{jia2023thinktwice} & CVPR'23 & 62.44 & 31.23 & 69.33  & 16.22 & 13.72 & 22.93 & 52.99 & \textbf{50.00} & 47.78 & 37.48 \\
DriveAdapter$^{*}$~\citep{jia2023driveadapter} & ICCV'23 & 64.22 & 33.08 & 70.22  & 16.01 & 14.55 & 22.61 & 54.04 & \textbf{50.00} & 50.45 & 38.33 \\
VAD~\citep{jiang2023vad} & ICCV'23 & 42.35 & 15.00 & 157.94 & 46.01 & 8.11  & 24.44 & 18.64 & 20.00 & 19.15 & 18.07 \\
GenAD~\citep{zheng2024genad} & ECCV'24 & 44.81 & 15.90 & -- & -- & -- & -- & -- & -- & -- & -- \\
DriveTransformer~\citep{jia2025drivetransformer} & ICLR'25 & 63.46 & 35.01 & 100.64 & 20.78 & 17.57 & 35.00 & 48.36 & 40.00 & 52.10 & 38.60 \\
SparseDrive~\citep{sun2025sparsedrive} & ICRA'25 & 44.54 & 16.71 & 170.21 & 48.63 & -- & -- & -- & -- & -- & -- \\
SimLingo~\citep{renz2025simlingo} & CVPR'25 & 86.02 & 67.27 & \textbf{259.23} & 33.67 & -- & -- & -- & -- & -- & -- \\
Hydra-NeXt~\citep{li2025hydranext} & ICCV'25 & 73.86 & 50.00 & 197.76 & 20.68 & 40.00 & 64.44 & 61.67 & \textbf{50.00} & 50.00 & 53.22 \\
HiP-AD~\citep{tang2025hipad} & ICCV'25 & 86.77 & 69.09 & 203.12 & 19.36 & 50.00 & \textbf{84.44} & \textbf{83.33} & 40.00 & \textbf{72.10} & 65.98 \\
MomAD (SD)~\citep{song2025momad} & CVPR'25 & 47.91 & 18.11 & 174.91 & 51.20 & 13.21 & 21.02 & 18.01 & 20.00 & 21.07 & 18.66 \\
FUMP~\citep{liu2025fump} & arXiv'25 & 45.67 & 16.36 & -- & -- & 12.50 & 24.44 & 20.00 & 21.50 & 19.15 & 19.51 \\
DIVER (SD)~\citep{song2026diver} & TPAMI'26 & 49.21 & 21.56 & 177.00 & 54.72 & 15.98 & 28.22 & 23.71 & 20.00 & 24.38 & 22.46 \\
GraphWorld (SD)~\citep{song2026graphworld} & arXiv'26 & 51.55 & 25.47 & 181.12 & \textbf{56.59} & 18.74 & 31.66 & 25.30 & 20.00 & 26.66 & 24.47 \\
GuideFlow~\citep{liu2025guideflow} & CVPR'26 & 75.21 & 51.36 & -- & -- & -- & -- & -- & -- & -- & -- \\
SparseDriveV2~\citep{sun2026sparsedrivev2} & ECCV'26 & \textbf{89.15} & \textbf{70.00} & 199.84 & 18.32 & \textbf{66.25} & 75.55 & 75.00 & \textbf{50.00} & 71.57 & \textbf{67.67} \\
\midrule
\rowcolor{red!5}
\textbf{MomWorld (Ours)}
& \multicolumn{1}{c}{--}
& 74.07
& 50.00
& 198.87
& 20.43
& 45.63
& 64.21
& 61.71
& \textbf{50.00}
& 54.10
& 55.07 \\
\bottomrule
\end{tabular}}
\end{table*}

\noindent\textbf{Bench2Drive.}
As shown in Table~\ref{tab_b2d}, MomWorld achieves a Driving Score of 74.07 and
a Success Rate of 50.00\%, together with an efficiency score of 198.87, a
comfort score of 20.43, and a mean ability score of 55.07. Compared with
Hydra-NeXt~\citep{li2025hydranext}, MomWorld improves Driving Score by 0.21
points and mean ability by 1.85 points, while matching its Success Rate and
maintaining similar efficiency and comfort. It also records higher Driving
Score, Success Rate, and mean ability than the related MomAD, DIVER, and
GraphWorld baselines~\citep{song2025momad,song2026diver,song2026graphworld}.
These results indicate more balanced route-level and interaction performance,
although HiP-AD and SparseDriveV2 retain higher Driving Score, Success Rate,
and mean ability~\citep{tang2025hipad,sun2026sparsedrivev2}.
We further evaluate robustness under turning-heavy motion, adversarial
interactions, and adverse-weather corruptions. Following DIVER and
GraphWorld~\citep{song2026diver,song2026graphworld}, all three experiments
report predicted collision rate at 1, 2, and 3 seconds. Turning-nuScenes also
reports L2 displacement where the original source provides it. Published
values are literature references rather than matched reruns. The shaded
MomWorld rows are reserved for evaluations using identical data manifests and
evaluator settings. The asterisk preserves the reimplementation mark used by
DIVER.

\begin{figure*}[t]
  \centering
  \includegraphics[width=\linewidth]{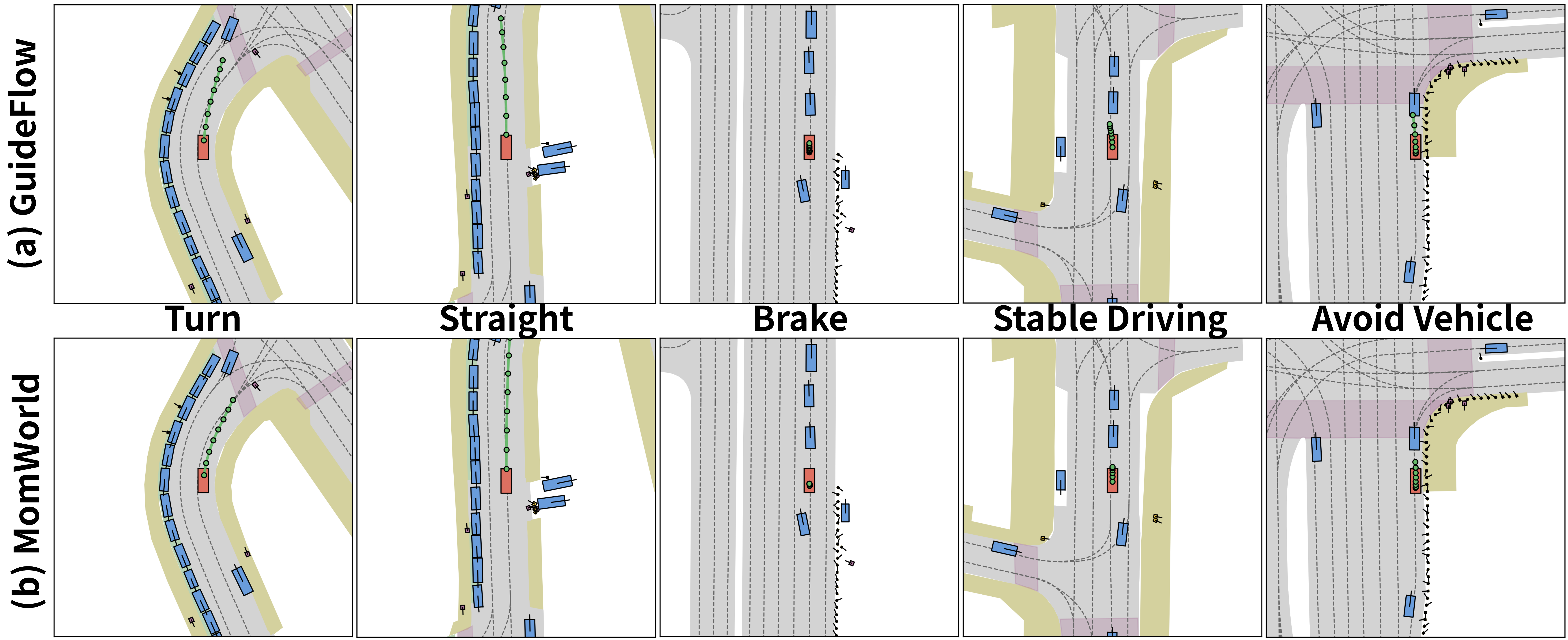}
  \caption{
  \textbf{Qualitative planning results on NAVSIM.}
  Comparison of GuideFlow~\citep{liu2025guideflow} and MomWorld across turning,
  straight driving, braking, stable driving, and vehicle avoidance. MomWorld
  produces smoother trajectories with improved roadway compliance and safer
  interactions.
  }
  \label{vis_planning_nav}
\end{figure*}

\begin{table*}[t]
\centering
\caption[]{Open-loop robustness on \textbf{Turning-nuScenes}~\citep{song2025momad}
and \textbf{Adv-nuSc}~\citep{xu2025challenger}. Turning-nuScenes additionally
reports average $L_2$ where available. Lower is better.}
\label{tab:app_nusc_shift_robustness}
\scriptsize
\renewcommand{\arraystretch}{0.98}
\setlength{\tabcolsep}{3.25mm}
\resizebox{\linewidth}{!}{%
\begin{tabular}{l c cccc @{\hspace{2.5mm}} cccc}
\toprule
\multirow{3}{*}{Method}
& \multicolumn{5}{c}{Turning-nuScenes}
& \multicolumn{4}{c}{Adv-nuSc} \\
\cmidrule(lr){2-6}\cmidrule(lr){7-10}
& \multirow{2}{*}{Avg. $L_2$ (m)$\downarrow$}
& \multicolumn{4}{c}{Col. Rate (\%)$\downarrow$}
& \multicolumn{4}{c}{Col. Rate (\%)$\downarrow$} \\
\cmidrule(lr){3-6}\cmidrule(lr){7-10}
& & 1s & 2s & 3s & Avg. & 1s & 2s & 3s & Avg. \\
\midrule
UniAD
& -- & -- & -- & -- & --
& 0.800 & 4.100 & 6.960 & 3.950 \\
VAD
& -- & -- & -- & -- & --
& 4.460 & 7.590 & 9.080 & 7.050 \\
SparseDrive
& 0.86 & 0.04 & 0.17 & 0.98 & 0.40
& 0.029 & 0.618 & 2.430 & 1.026 \\
DiffusionDrive$^{*}$
& -- & 0.03 & 0.14 & 0.85 & 0.34
& 0.068 & 1.299 & 3.646 & 1.671 \\
MomAD
& 0.76 & 0.03 & 0.13 & 0.79 & 0.32
& -- & -- & -- & -- \\
DIVER
& -- & 0.03 & \textbf{0.11} & 0.67 & 0.27
& 0.033 & 0.423 & 1.798 & 0.752 \\
GraphWorld
& -- & 0.03 & 0.12 & 0.72 & 0.28
& 0.028 & 0.420 & 1.780 & 0.742 \\
\midrule
MomWorld (Ours)
& \textbf{0.72} & \textbf{0.02} & \textbf{0.11} & \textbf{0.66}
& \textbf{0.26} & \textbf{0.027} & \textbf{0.418}
& \textbf{1.755} & \textbf{0.733} \\
\bottomrule
\end{tabular}}
\end{table*}

\begin{table*}[t]
\centering
\caption[]{Collision rate under three adverse-weather corruptions on
\textbf{nuScenes-C}~\citep{dong2023corruptions}. Lower is better.}
\label{tab:app_nusc_c}
\scriptsize
\renewcommand{\arraystretch}{0.78}
\setlength{\tabcolsep}{3.35mm}
\resizebox{\linewidth}{!}{%
\begin{tabular}{l ccc ccc ccc}
\toprule
\multirow{2}{*}{Method}
& \multicolumn{3}{c}{Snow}
& \multicolumn{3}{c}{Rain}
& \multicolumn{3}{c}{Fog} \\
\cmidrule(lr){2-4}\cmidrule(lr){5-7}\cmidrule(lr){8-10}
& 1s & 2s & 3s & 1s & 2s & 3s & 1s & 2s & 3s \\
\midrule
SparseDrive
& 0.13 & 0.27 & 0.50 & 0.11 & 0.27 & 0.55 & 0.14 & 0.36 & 0.58 \\
DiffusionDrive$^{*}$
& 0.09 & 0.24 & 0.39 & 0.07 & 0.18 & 0.35 & 0.06 & 0.18 & 0.30 \\
MomAD
& 0.08 & 0.16 & 0.30 & 0.06 & 0.17 & 0.31 & 0.06 & 0.19 & 0.32 \\
DIVER
& \textbf{0.07} & \textbf{0.13} & \textbf{0.25}
& \textbf{0.05} & 0.16 & \textbf{0.27}
& \textbf{0.04} & \textbf{0.16} & \textbf{0.25} \\
GraphWorld
& \textbf{0.07} & 0.15 & 0.28
& \textbf{0.05} & \textbf{0.15} & 0.29
& 0.05 & 0.17 & 0.29 \\
\midrule
\rowcolor{red!5}MomWorld (Ours)& \textbf{0.06} & \textbf{0.13} & \textbf{0.23}
& \textbf{0.04} & 0.15 & \textbf{0.25}
& \textbf{0.04} & \textbf{0.16} & \textbf{0.23} 
 \\
\bottomrule
\end{tabular}}
\end{table*}

\noindent\textbf{Turning-nuScenes.}
As shown in Table~\ref{tab:app_nusc_shift_robustness}, MomWorld achieves the
lowest average $L_2$ error of 0.72\,m and average collision rate of 0.26\%,
improving upon the 0.76\,m and 0.27\% references reported by MomAD and
DIVER~\citep{song2025momad,song2026diver}. It also matches the best 2-s
collision rate of 0.11\% and reduces the 3-s result from 0.67\% to 0.66\%.
These results support the effectiveness of momentum-aware rollout for stable
planning through high-curvature maneuvers.

\noindent\textbf{Adv-nuSc.}
On Adv-nuSc, MomWorld obtains collision rates of 0.027\%, 0.418\%, and 1.755\%
at 1, 2, and 3 seconds, respectively, with an average of 0.733\%. It improves
the corresponding GraphWorld references at every horizon and reduces the
average collision rate from 0.742\% to 0.733\%~\citep{song2026graphworld}.
The consistent gains under adversarial interactions indicate improved
adaptation when surrounding agents deviate from previously observed trends.

\noindent\textbf{nuScenes-C.}
Table~\ref{tab:app_nusc_c} shows that MomWorld is best or tied at every
evaluated horizon under snow, rain, and fog. At 3 seconds, it achieves collision
rates of 0.23\%, 0.25\%, and 0.23\%, reducing the strongest published references
of 0.25\%, 0.27\%, and 0.25\% by 0.02 percentage points in each condition
\citep{song2026diver,song2026graphworld}. The consistent performance across
weather corruptions demonstrates stronger robustness to degraded visual
observations.

\subsection{Additional Qualitative Results}
\label{app:Qualitative}
To complement the quantitative evaluation, we examine the planning behavior of
MomWorld on nuScenes and NAVSIM. The selected cases cover extended-horizon
maneuvers, dense traffic, changing road geometry, braking, and vehicle
avoidance. We present 6-second planning results on nuScenes and comparative
visualizations on NAVSIM.

\noindent\textbf{Long-Horizon Planning Visualization on nuScenes.}
Figure~\ref{nus_vis_fl} presents the 6-second planning behavior of MomWorld on
the nuScenes validation set. Across the four scenarios, MoLWM propagates
observed motion trends while adapting to predicted scene evolution, whereas
MoFlow converts these dynamics into bounded, horizon-aware corrections. The
resulting trajectories remain smooth and responsive throughout the extended
planning horizon.

\noindent\textbf{Planning Visualization on NAVSIM.}
Figure~\ref{vis_planning_nav} compares MomWorld with
GuideFlow~\citep{liu2025guideflow} across five representative scenarios.
MomWorld maintains smoother trajectories within drivable regions and adapts
more coherently to road geometry and surrounding vehicles. These examples
illustrate the benefits of latent momentum propagation and horizon-aware
residual refinement for stable, scene-responsive planning.

\bibliography{iclr2027_conference}
\bibliographystyle{iclr2027_conference}

\end{document}